\documentclass[10pt,twocolumn,letterpaper]{article}

\usepackage{3dv}
\usepackage{times}
\usepackage{epsfig}
\usepackage{graphicx}
\usepackage{amsmath}
\usepackage{amssymb}

\usepackage{bm}
\usepackage{mathtools}
\usepackage{subcaption}
\usepackage{xcolor}
\usepackage{multirow}
\usepackage[section]{placeins}
\usepackage{nicefrac}
\usepackage{overpic}
\usepackage[symbol]{footmisc}
\usepackage[titletoc]{appendix}

\DeclareMathOperator*{\argmin}{argmin}
\DeclareMathOperator*{\atantwo}{atan2}
\newcommand{\temporalVar}[1]{\sideset{^{t}}{}{\mathop{#1_{t-1}}}}
\newcommand{\temporalVarVar}[1]{\sideset{^{t - 1\hspace{-1pt}}}{}{\mathop{#1_{t-2}}}}

\newcommand{\temporalVarf}[1]{\sideset{^{t}}{^0_{t-1}}{\mathop{#1}}}
\newcommand{\temporalVarn}[1]{\sideset{^{t}}{^n_{t-1}}{\mathop{#1}}}

\newcommand{\approach}{\textit{TemporalLidarSeg }}
\newcommand{\approachq}{\textit{TemporalLidarSeg}}
\def\rot90{\rotatebox{90}}

\newcommand{\colBox}[1]{\tikz[baseline=(char.base)]{
            \node[shape=rectangle,draw=#1,inner sep=0.5pt, fill=#1, text=#1] (char) {T};}}

\definecolor{col_bicycle}{RGB}{245, 230, 100}
\definecolor{col_person}{RGB}{255,   125, 125}
\definecolor{col_sidewalk}{RGB}{75,0,75}
\definecolor{col_parking}{RGB}{255,150,255}
\definecolor{col_building}{RGB}{0,200,255}
\definecolor{col_other_veh}{RGB}{255,   190,   90}

\usepackage[pagebackref=true,breaklinks=true,letterpaper=true,colorlinks,bookmarks=false]{hyperref}
\usepackage{tikz}

\threedvfinalcopy % *** Uncomment this line for the final submission

\def\threedvPaperID{81} % *** Enter the 3DV Paper ID here

\ifthreedvfinal\fi

\begin{document}

%%%%%%%%% TITLE
\title{LiDAR-based Recurrent 3D Semantic Segmentation with Temporal Memory Alignment}

\author{Fabian Duerr\textsuperscript{\,1,\,2} \hspace{0.3cm} Mario Pfaller\textsuperscript{\,1} \hspace{0.3cm} Hendrik Weigel\textsuperscript{\,1} \hspace{0.3cm} Jürgen Beyerer\textsuperscript{\,2,\,3}
\and
\hspace{-0.4cm} \large \textsuperscript{1}AUDI AG \\
\large Ingolstadt, Germany \\
{\tt\small firstname.lastname@audi.de}
\and 
\hspace{-0.4cm} \large \textsuperscript{2} Vision and Fusion Lab \\
\large Karlsruhe Institute of Technology \\
\large Karlsruhe, Germany
\and
\hspace{-0.42cm} \large \textsuperscript{3} Fraunhofer IOSB\protect\footnote{Institute of Optronics, System Technologies and Image Exploitation IOSB Fraunhofer Center for Machine Learning} \\
\large Karlsruhe, Germany \\[-0.2ex]
{\tt\small juergen.beyerer} \\[-0.5ex] {\tt\small @iosb.fraunhofer.de}
}

\maketitle
\thispagestyle{empty}

%%%%%%%%% ABSTRACT
\begin{abstract}
Understanding and interpreting a 3d environment is a key challenge for autonomous vehicles. Semantic segmentation of 3d point clouds combines 3d information with semantics and thereby provides a valuable contribution to this task. In many real-world applications, point clouds are generated by lidar sensors in a consecutive fashion. Working with a time series instead of single and independent frames enables the exploitation of temporal information. %
We therefore propose a recurrent segmentation architecture (RNN), which takes a single range image frame as input and exploits recursively aggregated temporal information. An alignment strategy, which we call Temporal Memory Alignment, uses ego motion to temporally align the memory between consecutive frames in feature space. A Residual Network and ConvGRU are investigated for the memory update. %
We demonstrate the benefits of the presented approach on two large-scale datasets and compare it to several state-of-the-art methods. Our approach ranks first on the SemanticKITTI \cite{Behley2019SemanticKITTIAD} multiple scan benchmark and achieves state-of-the-art performance on the single scan benchmark. In addition, the evaluation shows that the exploitation of temporal information significantly improves segmentation results compared to a single frame approach.
\footnotetext[1]{Fraunhofer Institute of Optronics, System Technologies and Image Exploitation. Member of Fraunhofer Center for Machine Learning.}
\end{abstract}
%----------------------------------------------------------------------------------------------------------------------------------------------------------------------------------
%----------------------------------------------------------------------------------------------------------------------------------------------------------------------------------
%----------------------------------------------------------------------------------------------------------------------------------------------------------------------------------
%----------------------------------------------------------------------------------------------------------------------------------------------------------------------------------
\renewcommand*{\thefootnote}{\arabic{footnote}}
\section{Introduction}
Driven by the growing importance of autonomous driving and mobile agents, 3d scene understanding and interpretation has become more and more important. A key element for solving this challenge is semantic segmentation of sensor data, which enhances, e.g. images or 3d point clouds with valuable semantic information, by assigning a class label to every pixel or 3d point. \\
Semantic segmentation of 3d point clouds is particularly useful, because it combines 3d information with semantics. In order to exploit Convolutional Neural Networks (CNN) for this task, the representation of the raw point clouds has to be considered. Point based approaches \cite{Qi2017PointNetDL, Thomas2019} operate directly on the raw point clouds and apply specialized architectures and operations. Projection based methods \cite{milioto2019iros, Tchapmi2017} transform the point clouds into a convolution enabling regular space, e.g, 2d or  3d grid.  Representing the points using spherical coordinates and projecting them onto a range image is one efficient method of the latter category, which has shown promising results \cite{milioto2019iros, Xu2020}. It allows the application of established image segmentation architectures. \\
%##########################################################################################################################################
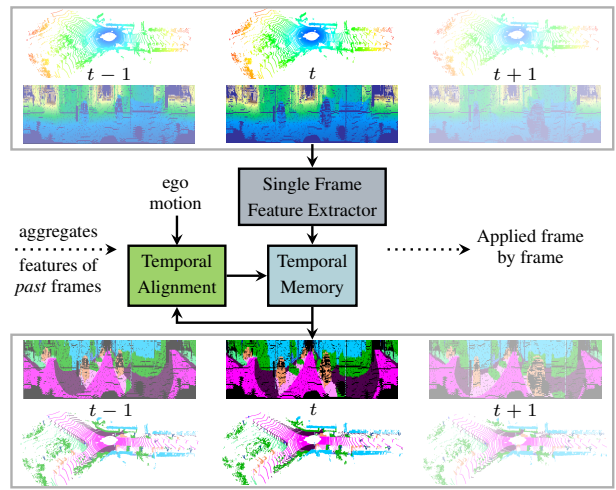
\begin{figure}[!t]
\begin{tikzpicture}[minimum width=0.5cm, minimum height=0.7cm, line width=0.9pt, >=stealth, bend angle=45, auto]
	\definecolor{lgreen}{RGB}{180,220,140}
	\definecolor{lgray}{RGB}{175,175,175}
    
	\definecolor{ncolB}{RGB}{173, 182, 191}	
	\definecolor{ncolC}{RGB}{179, 210, 217}
	\definecolor{ncolD}{RGB}{158, 210, 106}   
    
    % Input series
    \node[shape=rectangle,minimum height=0.4cm, inner sep=0pt, opacity=0.65] (inp0) at (1.4,4.2) {\includegraphics[scale=0.06]{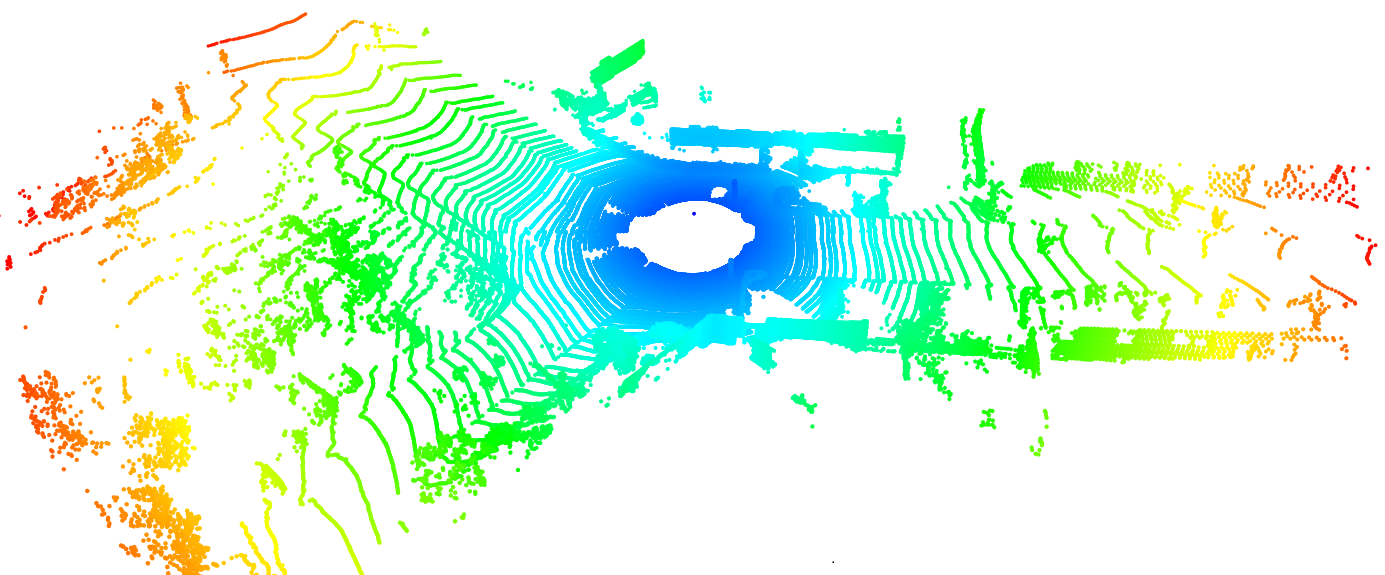}};
    \node[] (predTag0) at (1.4, 3.77) {\scriptsize $t-1$};
    \node[shape=rectangle,minimum height=0.4cm, inner sep=0pt, opacity=0.65] (inp0) at (1.4,3.25) {\includegraphics[scale=0.15]{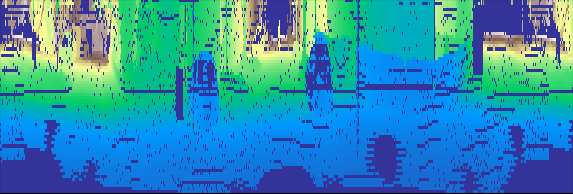}};   
    
    \node[shape=rectangle,minimum height=0.4cm, inner sep=0pt] (inp0) at (4.1,4.2) {\includegraphics[scale=0.06]{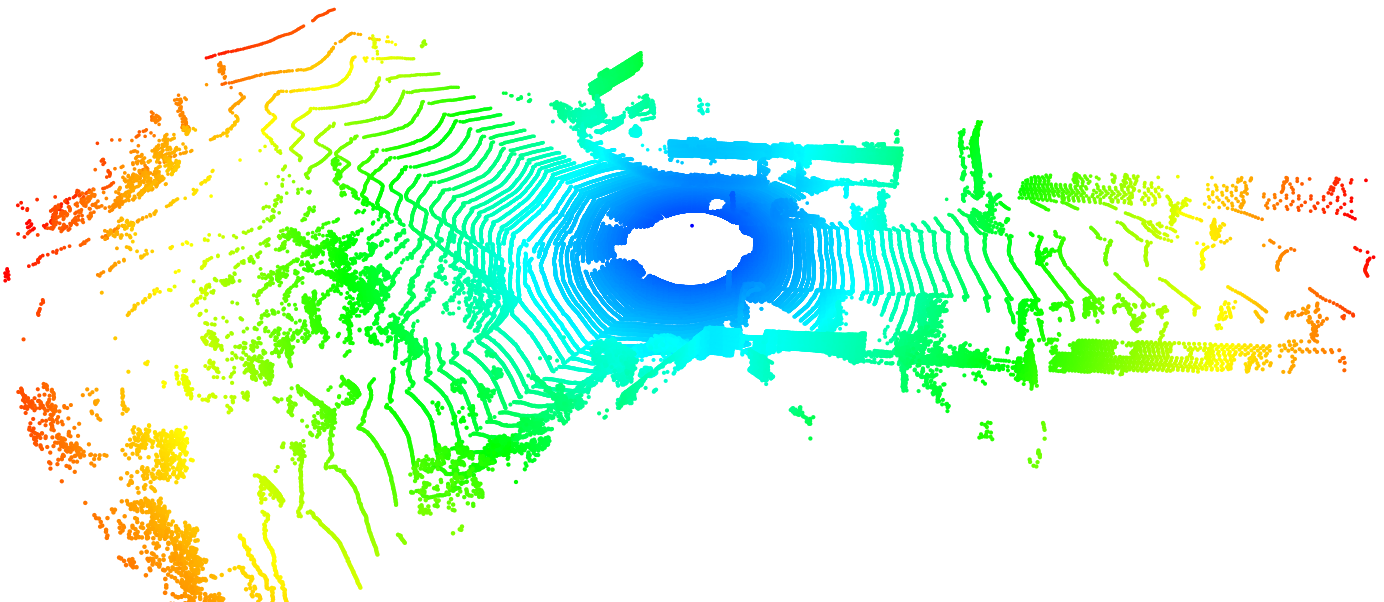}};
    \node[] (predTag0) at (4.1, 3.79) {\scriptsize $t$};
    \node[shape=rectangle,minimum height=0.4cm, inner sep=0pt] (inp1) at (4.1,3.25) {\includegraphics[scale=0.15]{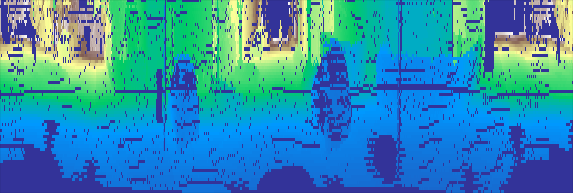}};

    \node[shape=rectangle,minimum height=0.4cm, inner sep=0pt, opacity=0.35] (inp0) at (6.8,4.2) {\includegraphics[scale=0.06]{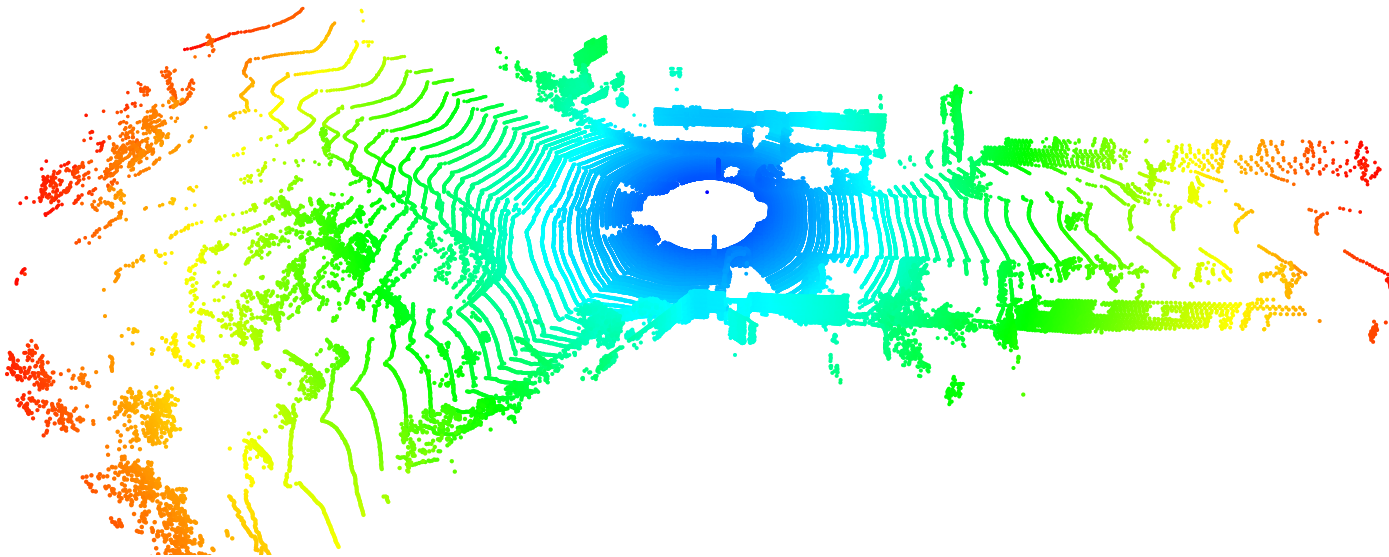}};;
    \node[] (predTag0) at (6.8, 3.77) {\scriptsize $t+1$};
    \node[shape=rectangle,minimum height=0.4cm, inner sep=0pt, opacity=0.35] (inp2) at (6.8,3.25) {\includegraphics[scale=0.15]{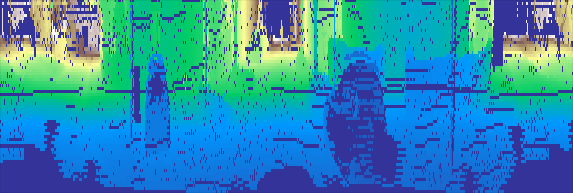}};
    
    \node[shape=rectangle,draw=lgray,opacity=1,xscale=16.0,yscale=2.7] (lidarbox) at (4.1,3.725) {};

	% Recurrent architecture
    \node[shape=rectangle,draw=black,fill=ncolB, align=center] (singleFrame) at (4.1,2.15) {\scriptsize Single Frame \\[-0.5ex] \scriptsize Feature Extractor};
    \node[shape=rectangle,draw=black,fill=ncolC, align=center] (temporalFusion) at (4.1,1.1) {\scriptsize Temporal \\[-0.5ex] \scriptsize Memory};
    \node[shape=rectangle,draw=black,fill=ncolD, align=center] (warping) at (2.3, 1.1) {\scriptsize Temporal \\[-0.5ex] \scriptsize Alignment};
    
    % Prediction series
    \node[shape=rectangle,minimum height=0.4cm, inner sep=0pt, opacity=0.625] (prediction0) at (1.4,-0.15) {\includegraphics[scale=0.15]{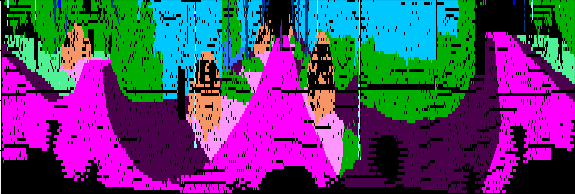}};
    \node[shape=rectangle,minimum height=0.4cm, inner sep=0pt, opacity=0.625] (pcPrad0) at (1.4,-1.2) {\includegraphics[scale=0.065]{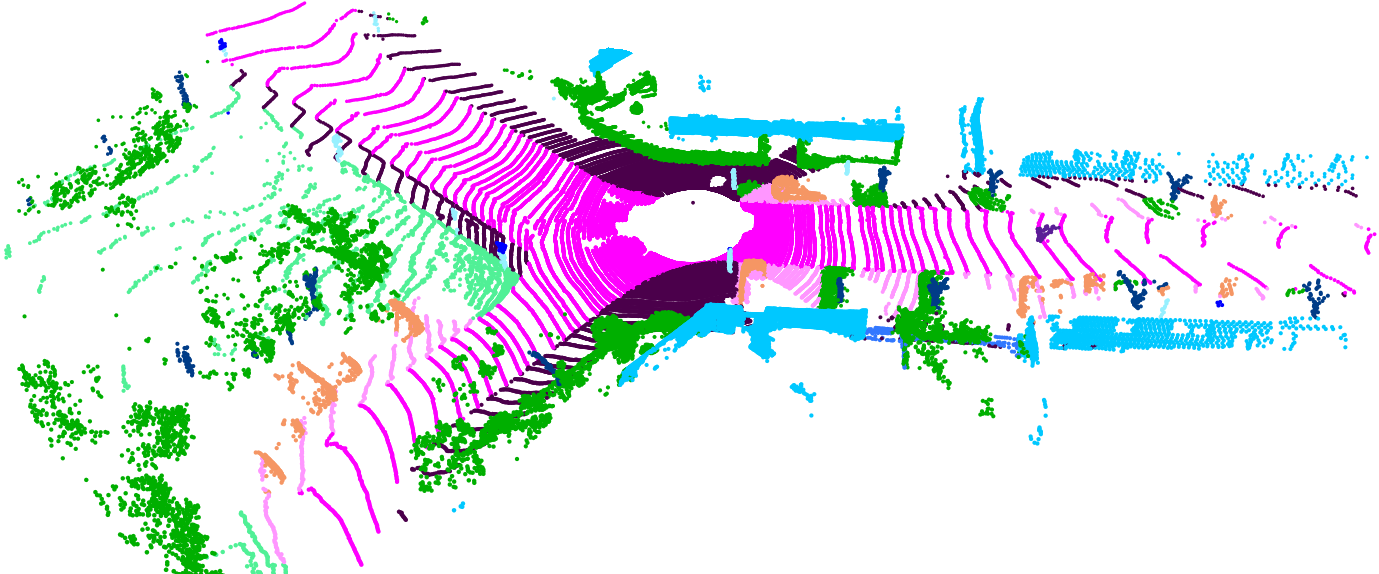}};
    \node[] (predTag0) at (1.425, -0.7) {\scriptsize $t-1$};
    
    \node[shape=rectangle,minimum height=0.4cm, inner sep=0pt] (prediction1) at (4.1,-0.15) {\includegraphics[scale=0.15]{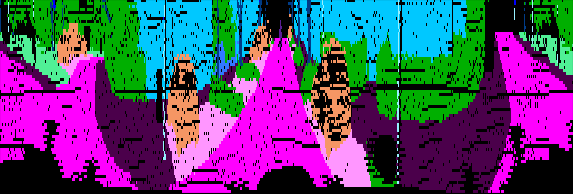}};
    \node[shape=rectangle,minimum height=0.4cm, inner sep=0pt] (pcPrad0) at (4.1,-1.2) {\includegraphics[scale=0.065]{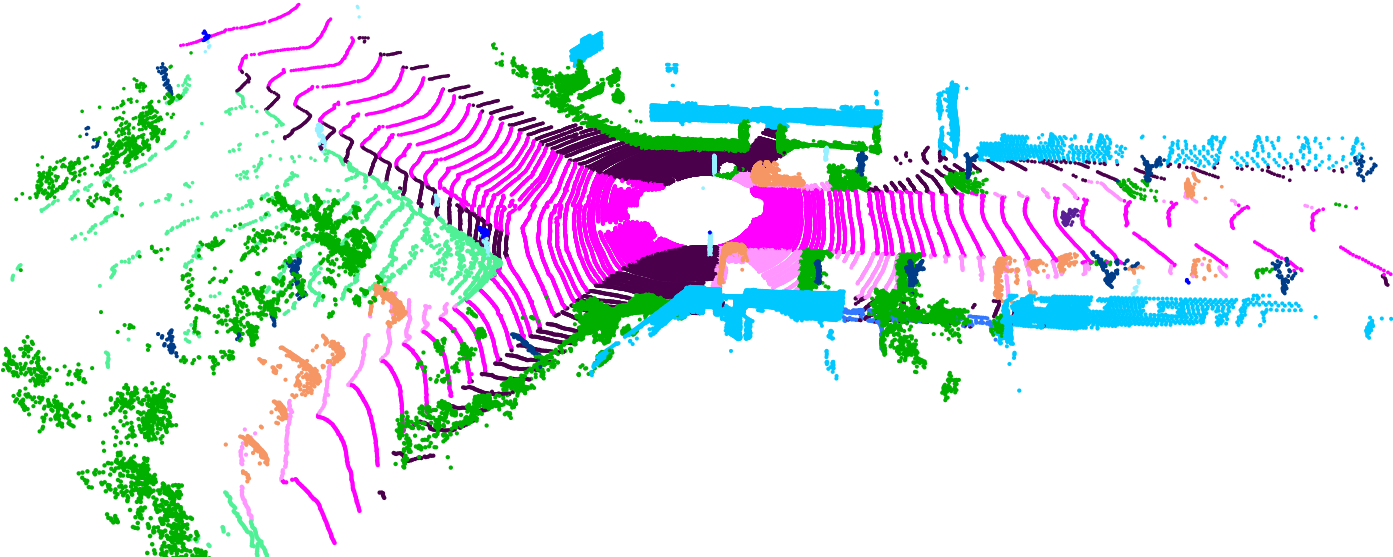}};
    \node[] (predTag0) at (4.115, -0.675) {\scriptsize $t$};  
	
    \node[shape=rectangle,minimum height=0.4cm, inner sep=0pt, opacity=0.35] (prediction2) at (6.8,-0.15) {\includegraphics[scale=0.15]{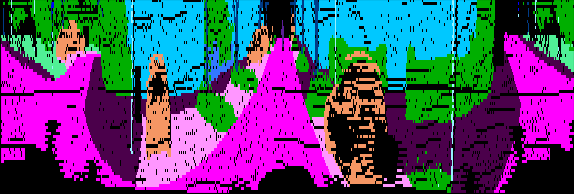}};
    \node[shape=rectangle,minimum height=0.4cm, inner sep=0pt, opacity=0.35] (pcPrad0) at (6.8,-1.2) {\includegraphics[scale=0.065]{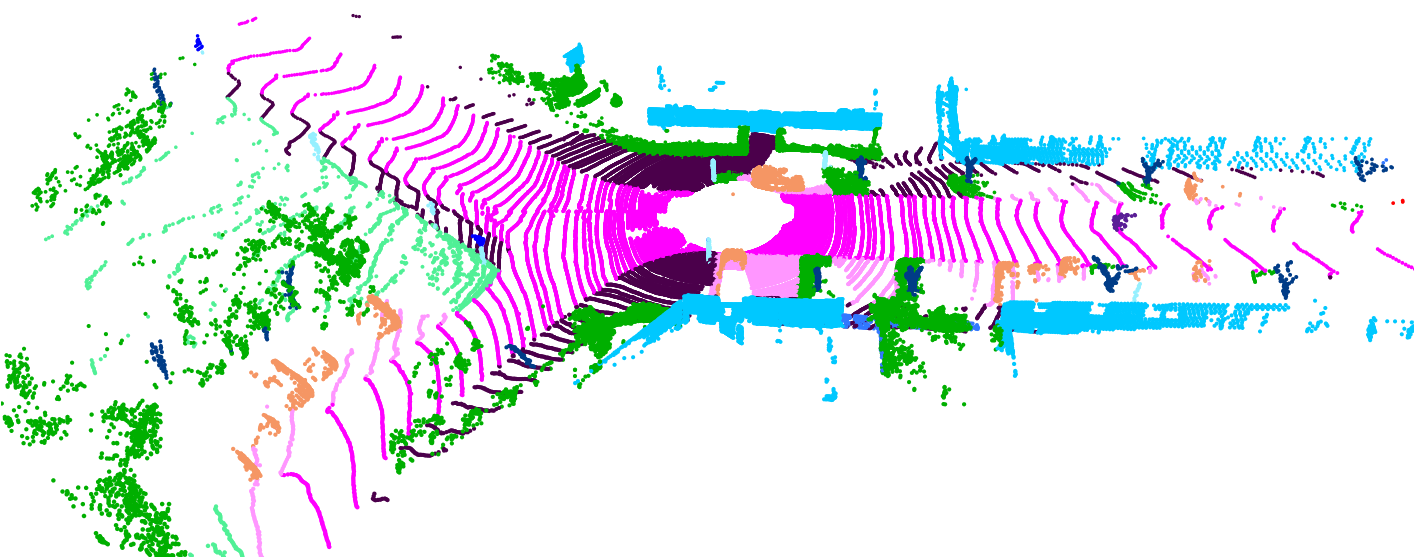}};
    \node[] (predTag0) at (6.8, -0.7) {\scriptsize $t+1$};   
	
    \node[shape=rectangle,draw=lgray,opacity=1,xscale=16.0,yscale=2.9] (lidarbox) at (4.1,-0.7) {};                
    
    % Edges
    \draw [->] (2.3,1.9) -- node[yshift=0.48cm, xshift=-0.5cm, align=center] {\scriptsize ego\\[-0.85ex] \scriptsize motion} (2.3,1.5);	
        
    \path [->](inp1) edge node {} (singleFrame);   
    \path [->](singleFrame) edge node {} (temporalFusion);
    \path [->](temporalFusion) edge node {} (prediction1);
    \path [->](warping) edge node {} (temporalFusion);
    \draw [->] (4.1, 0.475) -| (warping);       
    
    \draw [->, dotted] (5.1,1.45) -- node[yshift=-0.37cm, xshift=1.35cm, align=center] {\scriptsize Applied frame \\[-0.85ex] \scriptsize by frame} (6.2,1.45);	
    \draw [->, dotted] (0.15,1.45) -- node[yshift=-0.75cm, xshift=-0.1cm, align=center] {\scriptsize aggregates \\[0.2ex] \scriptsize features of \\[-0.9ex] \scriptsize \textit{past} frames} (1.5,1.45);	
\end{tikzpicture}
\caption{Proposed recurrent segmentation architecture, which is applied frame by frame to a time series of point clouds represented as range images. The approach exploits short term temporal dependencies to improve segmentation.}
\label{fig:motivation}
\vspace{-0.05cm}
\end{figure}%
%##########################################################################################################################################
3d semantic segmentation is usually considered as a single frame problem. However, lidar sensors, which record a vehicle's or robot's environment, provide a time series of point clouds, see Fig. \ref{fig:motivation}. Two frames adjacent in time share a lot of information, since the environment rarely changes drastically in that short amount of time. Therefore, previous frames still contain valuable information, which naturally diminish with increasing distance in time. As a result, especially short term temporal information or dependencies offer a huge potential to improve the segmentation results for the current frame. While it is very common in, e.g. natural language processing to rely on temporal dependencies and propagate information along a time series, it is rather uncommon for semantic segmentation of 3d point clouds. Here, we focus on real-time applications. Thus, the time series cannot contain future frames to minimize latency. \\%
In this work, we present a recurrent segmentation architecture for 3d semantic segmentation, illustrated in Fig. \ref{fig:motivation}. Features are extracted for every frame by a single frame feature extractor. By recursively updating a temporal memory with these features, temporal dependencies are exploited and a temporal memory alignment step ensures consistency between frames. Because of its recurrent nature, previous information can continuously and efficiently be reused. To summarize, our main contributions are:
\begin{itemize}
\item A recurrent segmentation architecture, which is applied frame by frame to a time series of point clouds with potentially unlimited length and exploits temporal dependencies to improve segmentation results.
\vspace{-0.05cm}
\item A Temporal Memory Alignment (TMA) strategy, to align the features of the temporal memory between adjacent frames directly in the range image feature space.
\vspace{-0.43cm}
\item Generalized range view projection for sensors with non-uniform laser distribution.
\end{itemize}%
%----------------------------------------------------------------------------------------------------------------------------------------------------------------------------------
%----------------------------------------------------------------------------------------------------------------------------------------------------------------------------------
%----------------------------------------------------------------------------------------------------------------------------------------------------------------------------------
%----------------------------------------------------------------------------------------------------------------------------------------------------------------------------------
\section{Related Work}
\subsection{3D Semantic Segmentation}
Driven by the growing importance of 3d lidar point cloud segmentation in robotics and autonomous driving, considerable progress has recently been achieved \cite{Hu2019,milioto2019iros,Thomas2019,Zhang2020}. This was fundamentally supported by an increasing number of publicly available indoor \cite{Armeni20163DSP} and outdoor lidar datasets \cite{Roynard2017, Hackel2017Semantic3DnetAN, Behley2019SemanticKITTIAD}. One crucial question, when addressing semantic segmentation of 3d point clouds with CNNs, is their input representation. Therefore, existing approaches can be grouped into two categories. \\
Point based methods operate directly on the raw point clouds without an initial transformation step. PointNet \cite{Qi2017PointNetDL} and its successor PointNet++ \cite{Qi2017PointNetDH} were one of the first approaches of this category, followed by a growing number of methods \cite{Hu2019,Li2018,Thomas2019,Wang2018DeepPC}. \\
Projection based methods transform the point clouds into a convolution enabling grid space and various projections have been proposed. Voxel-based approaches \cite{Choy2019, Tchapmi2017, Graham20183DSS} project the point clouds to a 3d voxel grid and apply 3d convolutions afterwards. While achieving convincing results, they suffer from heavy memory and computational costs. \cite{Rosu2019, Su2018} embed the point clouds into a sparse permutohedral lattice and apply sparse bilateral convolutions. Another possibility is either a  bird’s eye view \cite{Zhang2018EfficientCF, Zhang2020} or spherical projection. The latter results in a so called range image, which is also the foundation of the presented approach. SqueezeSeg \cite{Wu2017SqueezeSegCN} was one of the first methods using the range image for a segmentation task. Their goal was the segmentation of road objects, with an improved version released in \cite{Wu2018SqueezeSegV2IM}. Their third version SqueezeSegV3 \cite{Xu2020} targets full semantic segmentation and proposes Spatially-Adaptive Convolution (SAC) to counteract spatially-varying feature distributions in the range images. Another approach is RangeNet++ \cite{milioto2019iros}, which employs the DarkNet53 backbone \cite{Redmon2018YOLOv3AI} for semantic segmentation and specifically focuses on the backprojection of the predictions from range image space to 3d point clouds. LaserNets \cite{Meyer2019LaserNetAE} primary goal is 3d object detection, while one of their intermediate results is a semantic segmentation of the input. Their architecture is based on deep layer aggregation \cite{Yu2017DeepLA}. 3D-MiniNet \cite{Alonso2020} uses the range image for a fast neighbor search which is their foundation for a projection learning module. The learned projection is then fed to a Fully Convolutional Neural Network.
%----------------------------------------------------------------------------------------------------------------------------------------------------------------------------------
%----------------------------------------------------------------------------------------------------------------------------------------------------------------------------------
\subsection{Temporal 3D Semantic Segmentation}
The previously mentioned approaches treat the semantic segmentation task as a single frame problem. There are only few methods, which focus on the multi-frame task and the exploitation of temporal dependencies. \\%
The approach in \cite{Choy2019} uses 4d sparse tensors to incorporate previous frames and applies sparse operations to deal with the drawback of the mostly empty space. They explicitly don't use an RNN but convolutions for the temporal axis. One disadvantage is, that it doesn't scale well with the number of past frames used. SpSequenceNet \cite{ShiSpSequenceNetSS} uses a voxel-based representation and builds upon the backbone of \cite{Graham20183DSS}. It includes cross-frame global attention, an attention layer, which uses information from the last frame to highlight features of the current frame. Additionally, cross-frame local interpolation combines local information from previous and current frame. Generally, this approach only uses the last frame in addition to the current frame, which limits the amount of temporal information that can be learned and used. Both of the previous methods are not well suited for being applied continuously to a sequence of lidar point clouds, since they don't reuse features. \\%
For the task of 3d object detection, \cite{Yin2020} proposes an RNN based on an attentive spatiotemporal Transformer GRU, an extension to ConvGRU \cite{Ballas2015}. The task of this unit in the overall architecture is to aggregate spatiotemporal information and therefore, exploit temporal dependencies in point cloud sequences. The approach is designed for processing point cloud sequences in bird's eye view representation. \\
All mentioned approaches process a sequence of $n$ past frames for a prediction of the current frame, which is repeated when $\Delta t$ later the next frame arrives. The proposed approach however processes only the current frame, but reuses features of past frames, which have been computed in previous applications. Therefore, enabled by the temporal memory alignment, no re-computations for past frames are necessary and in combination with its recurrent structure, this allows for an efficient frame by frame application to a time series of point clouds of arbitrary length.%and still capturing temporal dependencies.
%----------------------------------------------------------------------------------------------------------------------------------------------------------------------------------
%----------------------------------------------------------------------------------------------------------------------------------------------------------------------------------
%----------------------------------------------------------------------------------------------------------------------------------------------------------------------------------
%----------------------------------------------------------------------------------------------------------------------------------------------------------------------------------
\section{Recurrent 3D Semantic Segmentation}
In this section, we describe the range image input representation and present our recurrent segmentation architecture and temporal training process.
%----------------------------------------------------------------------------------------------------------------------------------------------------------------------------------
\subsection{Adaptive Range Image} \label{ssec:rangeImage}
Lidar sensors usually observe their environment with a set of vertically stacked lasers, spinning around their vertical axis. Therefore, a native way to specify a measurements position in 3d space are spherical coordinates \(\left(\theta, \phi, r\right)\), with elevation \(\theta\), azimuth \({\phi}\) and measured distance \({r}\), instead of cartesian coordinates \(\bm{p}=\left(x, y, z\right)\). This may require a transformation of $\bm{p}$ into spherical coordinates, but enables a spherical projecting of the 3d points onto a 2d range image. Its resolution \(\left(h, w\right)\) is defined by the lidar's intrinsic properties. A natural projection strategy to  image coordinates \((u, v)\), is a simple discretization \cite{milioto2019iros, Xu2020} 
\begin{equation} \label{eq:simple_proj}
\begin{pmatrix}
u \\[0.4em] v
\end{pmatrix} = 
\begin{pmatrix}
\left\lfloor\left(1 - \left(\theta + f_{\text{up}}\right)f^{-1}\right) \cdot h \right\rfloor \\[0.7em]
\left\lfloor0.5\cdot\left(1 - \phi\cdot\pi^{-1}\right)\cdot w\right\rfloor 
\end{pmatrix},
\end{equation}%
with the lidar's field of view $f = f_{\text{up}} - f_{\text{down}}$. A challenge arises, if the lidar's vertically stacked lasers are non-uniformly distributed, like the sensor used for PandaSet\footnote{\url{https://pandaset.org/}}, leading to non-uniformly distributed elevation angles. Using the mapping in Eq. \ref{eq:simple_proj} leads to many collisions.  As a result, more than one point is mapped to the same range image pixel. This implies not only a loss of information but also missing predictions for the shadowed points. The latter isn't an issue for object detection, for semantic segmentation however, it has to be considered. To reduce the number of collisions, we propose an adapted projection strategy. Every row $l \in \left[0, h-1\right]$ in the range image corresponds to a laser in the vertical stack and every laser has an assigned elevation angle $\overline{\theta}_l$, due to design. To deal with the non-uniformity, the row with the closed matching elevation angle is assigned:%
\begin{equation} \label{eq:ext_proj}
u = \argmin_{0 \leq l < h}(\left|\overline{\theta}_l - \theta\right|).
\end{equation}%
Based on the computed image coordinates, $r$, $x$, $y$ and $z$, as wells as the remission $e$ and an occupancy flag, are mapped to a $6 \times h \times w$ range image. \\%
Ego motion and uncertainty of the angles can also lead to mapping collisions, which can't be prevented by the proposed mapping. Therefore, a post-processing step based on the labeled points is still required to compute class labels for the shadowed points. Following \cite{milioto2019iros} we use a majority voting of the k-nearest neighbors to determine the missing labels. The neighbor candidates for a 3d point are the image pixels in a 5$\times$5 window around the pixel corresponding to the 3d point. The k-nearest neighbors are then selected based on the range $r$. Further details can be found in \cite{milioto2019iros}.
%----------------------------------------------------------------------------------------------------------------------------------------------------------------------------------
%----------------------------------------------------------------------------------------------------------------------------------------------------------------------------------
\subsection{Temporal Memory Alignment}
%##########################################################################################################################################
\begin{figure}[!t]
\begin{tikzpicture}[minimum width=0.5cm, minimum height=0.7cm, line width=0.9pt, >=stealth, bend angle=45, auto]
	\definecolor{ncolD}{RGB}{158, 210, 106} 
	\definecolor{lblue}{RGB}{164,194,244}
	\definecolor{lgray}{RGB}{175,175,175}
	\definecolor{dgreen}{RGB}{29,148,65}
	\definecolor{dblue}{RGB}{77,116,168}
    
    % Nodes Fusion
    \node[shape=rectangle,draw=white, inner sep=0.0] (pt-1) at (0.1, 2.05) {\small $\bm{P}_{t-1}$};  
    \node[shape=rectangle,draw=white] (Tt-1) at (1.1, 2.6) {\small $\bm{T}_{t-1}$}; 
    \node[shape=rectangle,draw=white] (Tt) at (1.85, 2.6) {\small $\bm{T}_t$};      
    \node[shape=rectangle,draw=white] (pt) at (2.8, 2.05) {\small $\temporalVar{\bm{P}}$};   
    \node[shape=rectangle,draw=white] (u_v) at (5.63, 2.05) {\small $\temporalVar{\bm{u}}$};  
    \node[shape=rectangle,draw=white] (u_v) at (5.63, 1.55) {\small $\temporalVar{\bm{v}}$}; 
    \node[shape=rectangle,draw=white] (Ft-1) at (7.0, 3.2) {\small $\bm{H}_{t-1}$}; 
    \node[shape=rectangle,draw=white, inner sep=0pt, minimum height=0.0cm] (Ft) at (7, 0.725) {\small $\temporalVar{\bm{H}}$}; 
    \node[shape=rectangle,draw=black,fill=ncolD] (transform) at (1.4,1.8) {\footnotesize Transform (\ref{eq:transformation})};
    \node[shape=rectangle,draw=black,fill=ncolD] (projection) at (4.25,1.8) {\footnotesize Projection (\ref{eq:projection})};
    \node[shape=rectangle,draw=black,fill=ncolD] (warping) at (7.0,1.8) {\footnotesize Warping (\ref{eq:rearrange})};
    \node[shape=rectangle,draw=lgray,opacity=1,xscale=16.5,yscale=2.5] (lidarbox) at (3.8,2.0) {};
        
    \draw [->] (-0.2,1.8) -- (0.5,1.8);	
    \draw [->] (0.9,2.47) -- (0.9,2.15);	
    \draw [->] (1.8,2.47) -- (1.8,2.15);	    
    \path [->](transform) edge node {} (projection);
    \path [->](projection) edge node {} (warping);
    \draw [->] (7.0,3.0) -- (7.0,2.15);	
    \draw [->] (7.0,1.45) -- (7.0,0.9);		
\end{tikzpicture}
\vspace{-0.575cm}
\caption{\textbf{Temporal Memory Alignment (TMA)}. Based on the ego motion $\bm{T}_{t-1} \rightarrow \bm{T}_t$, the point cloud $\bm{P}_{t-1}$ of the last frame is transformed and projected. Hence, the temporal memory can be rearranged using $\temporalVar{\bm{u}}$ and $\temporalVar{\bm{v}}$.}
\label{fig:warping}
\vspace{-0.1cm}
\end{figure}
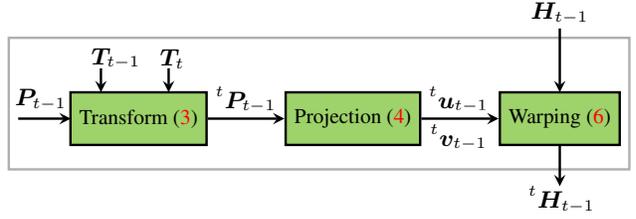%
%##########################################################################################################################################
%##########################################################################################################################################
\begin{figure*}[!t]
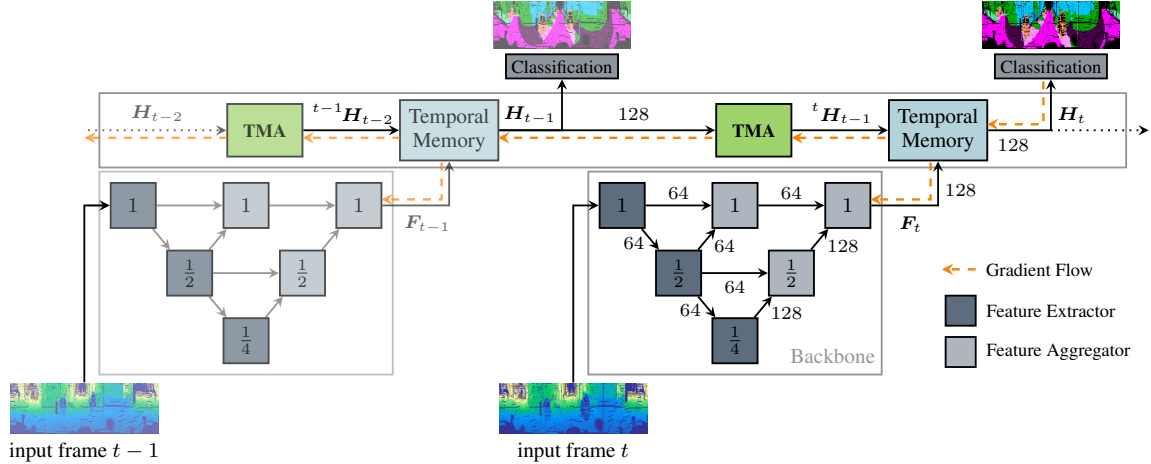

\centering
\begin{tikzpicture}[minimum width=0.6cm, minimum height=0.6cm,line width=0.7pt, >=stealth, auto]
	\definecolor{cgrey}{RGB}{142,149,157}
	\definecolor{cblue}{RGB}{95,151,177}
	\definecolor{cgreen}{RGB}{123,171,116}	
	
	\definecolor{lgray}{RGB}{150,150,150}
	\definecolor{dgreen}{RGB}{29,148,65}
	\definecolor{dblue}{RGB}{77,116,168}
	
	\definecolor{ncolA}{RGB}{93, 109, 125}
	\definecolor{ncolB}{RGB}{173, 182, 191}
	\definecolor{ncolC}{RGB}{179, 210, 217}
	\definecolor{ncolD}{RGB}{158, 210, 106}  
	
	%
    % Nodes backbone t-1
    \node[shape=rectangle,draw=black,fill=ncolA, opacity=0.7] (i1ap) at (0,1.8) {\footnotesize $1$};
    \node[shape=rectangle,draw=black,fill=ncolA, opacity=0.7] (i2ap) at (0.75,0.9) {\footnotesize $\frac{1}{2}$};
    \node[shape=rectangle,draw=black,fill=ncolA, opacity=0.7] (i3ap) at (1.5,0) {\footnotesize $\frac{1}{4}$};
    \node[shape=rectangle,draw=black,fill=ncolB, opacity=0.7] (i1bp) at (1.5,1.8) {\footnotesize $1$};
    \node[shape=rectangle,draw=black,fill=ncolB, opacity=0.7] (i2bp) at (2.25,0.9) {\footnotesize $\frac{1}{2}$};  
    \node[shape=rectangle,draw=black,fill=ncolB, opacity=0.7] (i1cp) at (3.0,1.8) {\footnotesize $1$};
    \node[shape=rectangle,draw=lgray,opacity=0.6,xscale=6.5, yscale=4.5] (lidarbox) at (1.5,0.9) {};
    \node[shape=rectangle,minimum height=0.4cm, inner sep=0pt, opacity=0.7] (inputp) at (-0.65,-0.9) {\includegraphics[scale=0.13]{images/range_t-1.png}};
    \node[inner sep=0pt] (frame_t-1) at (-0.65, -1.475) {\footnotesize input frame $t-1$}; 
	%
	% Nodes backbone t
    \node[shape=rectangle,draw=black,fill=ncolA] (i1a) at (6.5,1.8) {\footnotesize $1$};
    \node[shape=rectangle,draw=black,fill=ncolA] (i2a) at (7.25,0.9) {\footnotesize $\frac{1}{2}$};
    \node[shape=rectangle,draw=black,fill=ncolA] (i3a) at (8.0,0) {\footnotesize $\frac{1}{4}$};
    \node[shape=rectangle,draw=black,fill=ncolB] (i1b) at (8.0,1.8) {\footnotesize $1$};
    \node[shape=rectangle,draw=black,fill=ncolB] (i2b) at (8.75,0.9) {\footnotesize $\frac{1}{2}$};  
    \node[shape=rectangle,draw=black,fill=ncolB] (i1c) at (9.5,1.8) {\footnotesize $1$}; 
    \node[shape=rectangle,draw=lgray, minimum width=3.9cm, minimum height=2.7cm] (lidarbox) at (8.0,0.9) {}; 
    \node[shape=rectangle, text=lgray] (labelBackbone) at (9.3,-0.2) {\footnotesize Backbone}; 
   	\node[shape=rectangle,minimum height=0.4cm, inner sep=0pt] (input) at (5.85,-0.9) {\includegraphics[scale=0.13]{images/range_t.png}};
    \node[inner sep=0pt] (frame_t) at (5.85, -1.475) {\footnotesize input frame $t$}; 
	%
	% Edges backbone t-1 	
	\draw [->] (inputp) |- (i1ap);
	\draw [->, opacity=0.4] (-0.6, 1.8) -- (i1ap);
    \path [->, opacity=0.4](i1ap) edge node {} (i2ap);
    \path [->, opacity=0.4](i2ap) edge node {} (i3ap);
    \path [->, opacity=0.4](i1ap) edge node {} (i1bp);
    \path [->, opacity=0.4](i1bp) edge node {} (i1cp);
    \path [->, opacity=0.4](i2ap) edge node {} (i1bp);
    \path [->, opacity=0.4](i2ap) edge node {} (i2bp);
    \path [->, opacity=0.4](i3ap) edge node {} (i2bp);
    \path [->, opacity=0.4](i2bp) edge node {} (i1cp);	
	%
	% Edges backbone t
	\draw [->] (input) |- (i1a);
    \path [->](i1a) edge node[xshift=-0.53cm, yshift=-0.36cm] {\scriptsize $64$} (i2a);
    \path [->](i2a) edge node[xshift=-0.53cm, yshift=-0.36cm] {\scriptsize $64$} (i3a);
    \path [->](i1a) edge node[yshift=-0.15cm] {\scriptsize $64$} (i1b);
    \path [->](i1b) edge node[yshift=-0.15cm] {\scriptsize $64$} (i1c);
    \path [->](i2a) edge node[xshift=0.55cm, yshift=-0.4cm] {\scriptsize $64$} (i1b);
    \path [->](i2a) edge node[yshift=-0.5cm] {\scriptsize $64$} (i2b);
    \path [->](i3a) edge node[xshift=0.65cm, yshift=-0.4cm] {\scriptsize $128$} (i2b);
    \path [->](i2b) edge node[xshift=0.65cm, yshift=-0.4cm] {\scriptsize $128$} (i1c);
    %
    % Memory node t-1
    \node[shape=rectangle,draw=black, fill=ncolD, opacity=0.7, minimum height=0.7cm, minimum width =1.0cm] (tfap) at (1.75, 2.8) {\scriptsize \textbf{TMA}};
    \node[shape=rectangle,draw=black,fill=ncolC, opacity=0.7, minimum height=0.8cm, align=center] (fusionp) at (4.2,2.8) {\footnotesize Temporal \\[-0.3em] \footnotesize Memory};
	\node[shape=rectangle,draw=black,fill=cgrey, minimum height=0.2cm, inner sep = 0.075cm] (clsp) at (5.7,3.65) {\scriptsize Classification};
    \node[shape=rectangle,minimum height=0.4cm, inner sep=0pt, opacity=0.6] (predictionp) at (5.7,4.2) {\includegraphics[scale=0.12]{images/target_t-1.png}};    
    %
    % Memory node t
    \node[shape=rectangle,draw=black, fill=ncolD, minimum height=0.7cm, minimum width =1.0cm] (tfa) at (8.25, 2.8) {\scriptsize \textbf{TMA}};
    \node[shape=rectangle,draw=black,fill=ncolC, minimum height=0.8cm, align=center] (fusion) at (10.7,2.8) {\footnotesize Temporal \\[-0.3em] \footnotesize Memory};
    \node[shape=rectangle,draw=black,fill=cgrey, minimum height=0.2cm, inner sep = 0.075cm] (cls) at (12.2,3.65) {\scriptsize Classification};
    \node[shape=rectangle,minimum height=0.4cm, inner sep=0pt] (prediction1) at (12.2,4.2) {\includegraphics[scale=0.12]{images/target_t.png}};    %
    \node[shape=rectangle,draw=lgray,xscale=22.75, yscale=1.65] (lidarbox) at (6.375,2.8) {};
	%
	% Edges memory node t-1
	\draw [dotted, ->, opacity=0.6] (-0.6, 2.8) -- node[yshift=-0.1cm]{\scriptsize $\bm{H}_{t-2}$}(tfap);	
	\path [->](tfap) edge node[yshift=-0.1cm, xshift=0.0cm]{\scriptsize $\temporalVarVar{\bm{H}}$} (fusionp); 
	\draw [->, opacity=0.6] (i1cp) -| node[yshift=-0.25cm, xshift=0.15cm]{\scriptsize $\bm{F}_{t-1}$}(fusionp);
	\draw [->] (fusionp) -| node[xshift=0.05cm, yshift=0.2cm]{\scriptsize $\bm{H}_{t-1}$} (clsp);	
	%
	% Edges memory node t
	\path [->](fusionp) edge node[yshift=-0.1cm, xshift=0.35cm]{\scriptsize $128$} (tfa);	
	\path [->](tfa) edge node[yshift=-0.1cm, xshift=0.0cm]{\scriptsize $\temporalVar{\bm{H}}$} (fusion);	
	\draw [->] (i1c) -| node[yshift=-0.25cm, xshift=-0.05cm]{\scriptsize $\bm{F}_t$}(fusion);
	\draw [->] (fusion) -| node[yshift=0.2cm, xshift=0.6cm]{\scriptsize $\bm{H}_t$} (cls); 
	\draw [dotted, ->] (12.2, 2.8) -- (13.5, 2.8);	
	\node[] (label1) at (11.7,2.6) {\scriptsize $128$};
	\node[] (label2) at (11.0,2.0) {\scriptsize $128$};
	
	% gradient
	\draw [->, dashed, draw=orange, line width=0.9pt] (12.1, 3.45) |- (11.35, 2.88);
	\draw [->, dashed, draw=orange, line width=0.9pt] (10.6, 2.37) |- (9.8, 1.88);
	\draw [->, dashed, draw=orange, line width=0.9pt] (10.0, 2.7) |- (8.75, 2.7);
	\draw [->, dashed, draw=orange, line width=0.9pt] (7.7, 2.7) |- (4.85, 2.7);
	\draw [->, dashed, draw=orange, line width=0.9pt, opacity=0.7] (4.1, 2.37) |- (3.3, 1.88);
	\draw [->, dashed, draw=orange, line width=0.9pt, opacity=0.7] (3.55, 2.7) |- (2.25, 2.7);
	\draw [->, dashed, draw=orange, line width=0.9pt, opacity=0.7] (1.25, 2.7) |- (-.65, 2.7);
	
	% legend
	\node[shape=rectangle, minimum width=0.4cm, minimum height=0.4cm, draw=black,fill=ncolA, label=right:{\scriptsize  Feature Extractor}] (i3ap) at (11.0,0.4) {};
    \node[shape=rectangle, minimum width=0.4cm, minimum height=0.4cm, draw=black,fill=ncolB, label=right:{\scriptsize  Feature Aggregator}] (i1bp) at (11.0,-0.15) {};
    \draw [->, dashed, draw=orange, line width=0.9pt] (11.2, 0.95) -- node[yshift=0.31cm, xshift=1.075cm]{\scriptsize Gradient Flow} (10.75, 0.95);	
\end{tikzpicture}
\vspace{-0.25cm}
\caption{Overview of the recurrent fully convolutional architecture and its components, unrolled for two time steps. The edges are labeled with the number of feature channels and the Feature Extractors and Aggregator \cite{Meyer2019LaserNetAE} with the horizontal size ratio of their output feature maps with regard to the original network input. The vertical resolution stays constant.}
\label{fig:fusionArch}
\vspace{-0.075cm}
\end{figure*}%
%##########################################################################################################################################
For an efficient reuse of features of previous frames, the features need to be temporally aligned in the range image space. The alignment is necessary because the sensor reference system usually moves between two frames as a result of the ego motion. While camera image features are often aligned using optical flow \cite{Jain2018, Zhu2016, Gadde2017}, we propose a different alignment strategy. Despite working with 2d range images, we can exploit the cartesian position $\bm{p}$ of the points and the ego motion. In the first step, see Fig. \ref{fig:warping}, all points $\bm{P}_{t-1} = \left(\bm{p}^0_{t-1}, ...,  \bm{p}^n_{t-1}\right) = \left(\bm{x}_{t-1},\bm{y}_{t-1},\bm{z}_{t-1}\right)^T$  of the last frame, which were recorded relative to the last sensor pose $\bm{T}_{t-1}$ are transformed to the current frame at sensor pose $\bm{T}_{t}$:
\begin{equation} \label{eq:transformation}
\temporalVar{\bm{P}} = \bm{T}_{t}^{-1} \cdot  \bm{T}_{t-1} \cdot \bm{P}_{t-1},
\end{equation}%
using homogeneous coordinates. Based on this, the elevation and azimuth angle, and thereby the image coordinates, of the transformed spherical coordinates are given by
\begin{equation} \label{eq:projection}
\begin{gathered}
\temporalVar{\bm{\theta}} = \arcsin\left(\frac{\temporalVar{\bm{z}}}{\temporalVar{\bm{r}}}\right) \xRightarrow[]{\text{Eq.}\,\ref{eq:ext_proj}\,\,} \temporalVar{\bm{u}} \\ 
\temporalVar{\bm{\phi}} = -\atantwo\left(\temporalVar{\bm{y}}, \temporalVar{\bm{x}}\right) \xRightarrow[]{\text{Eq.}\,\ref{eq:simple_proj}\,\,} \temporalVar{\bm{v}} \\ 
\text{with} \temporalVar{\bm{r}} = \left(\left\|\temporalVarf{\bm{p}}\right\|, ..., \left\|\temporalVarn{\bm{p}}\right\|\right)^{T},
\end{gathered}
\end{equation}%
where $\atantwo$, $\arcsin$ and division are applied element-wise. The transformed image coordinates $\temporalVar{\bm{u}}$ and $\temporalVar{\bm{v}}$ describe the location of the measurements and corresponding features of the last frame in the current range image frame. With an index operation defined as 
\begin{equation} 
\bm{f}_{u, v} \leftarrow \bm{H} \left[u, v\right], \quad \bm{f}_{u, v} \in \mathbb{R}^{c},
\end{equation}%
for a channel dimension $c$ and arbitrary image coordinates $u$ and $v$, the feature vectors of a temporal memory $\bm{H}$ can be rearranged by:
\begin{equation} \label{eq:rearrange}
\begin{gathered}
\temporalVar{\bm{H}}\left[\temporalVar{u}, \temporalVar{v}\right]  \leftarrow  \bm{H}_{t-1}\left[u_{t-1}, v_{t-1}\right] \\
\forall \temporalVar{u}, \temporalVar{v} \in \temporalVar{\bm{u}}, \temporalVar{\bm{v}}. 
\end{gathered}
\end{equation}%
Transformed image coordinates outside the field of view will be ignored. Entries in $\temporalVar{\bm{H}}$ not assigned a feature vector from $\bm{H}_{t-1}$ are set to zero. The overall pipeline for aligning the temporal memory is illustrated in Fig. \ref{fig:warping}. The presented equations only consider the transformation between two adjacent frames. This is sufficient for the presented approach, because of its recursive nature. Nevertheless, the presented relations are straightforwardly generalizable to the temporal alignment of frames or features with arbitrary temporal distance. 
%----------------------------------------------------------------------------------------------------------------------------------------------------------------------------------
%----------------------------------------------------------------------------------------------------------------------------------------------------------------------------------
\subsection{Fusion Architecture}
\begin{figure}[!t]
\centering
\begin{tikzpicture}[minimum width=0.6cm, minimum height=0.6cm,line width=0.7pt, >=stealth, auto]
%	\definecolor{cgreen}{RGB}{123,171,116}	

	\definecolor{ncolC}{RGB}{179, 210, 217}
	\definecolor{ncolD}{RGB}{140, 160, 194}  

	% Nodes fusion module
    \node[shape=rectangle,draw=black,fill=ncolC, fill opacity=1.0, minimum height=1.8cm, minimum width =5.0cm] (fusionmodule) at (2.75, 0.985) {};
    
    \node[shape=circle,draw=black, minimum height=0.4cm, minimum width =0.4cm, inner sep=1pt] (concat) at (0.75, 1.0) {\footnotesize c};    
    \node[shape=rectangle,draw=black,fill=black, minimum height=1.0cm, minimum width =0.15cm, inner sep=0, label={[xshift=0.3cm, yshift=-1.5cm]:\scriptsize  $1$x$1$-conv + BN}] (conv1) at (1.5, 1.0) {};
    \node[shape=rectangle,draw=black,fill=black, minimum height=1.0cm, minimum width =0.15cm, inner sep=0] (bn) at (2.0, 1.0) {};
    
    \node[shape=rectangle,draw=black, fill=ncolD, minimum height=1.0cm, minimum width=0.275cm, inner sep=3pt] (resnet0) at (3.2, 1.0) {};
    \node[shape=rectangle,draw=black, fill=ncolD, minimum height=1.0cm, minimum width=0.275cm, inner sep=3pt, label={[xshift=0.275cm, yshift=-1.5cm]:\scriptsize  Residual Units}] (resnet1) at (3.7, 1.0) {};
    \node[shape=rectangle,draw=black, fill=ncolD, minimum height=1.0cm, minimum width=0.275cm, inner sep=3pt] (resnet2) at (4.2, 1.0) {};
    \node[shape=rectangle,draw=black, fill=ncolD, minimum height=1.0cm, minimum width=0.275cm, inner sep=3pt] (resnet3) at (4.7, 1.0) {};
	%
	% Edges fusion module	
	\draw [->] (0.0, 0.5) -| node[xshift=-0.6cm, yshift=0cm]{\footnotesize $\bm{F}_t$} (concat);	
	\draw [->] (0.0, 1.5) -| node[xshift=-1.8cm, yshift=0cm]{\footnotesize $\temporalVar{\bm{H}}$} (concat);	
	
	\draw [->] (concat) -- (conv1);
	\draw [->] (conv1) -- (bn);
	\draw [->] (bn) -- (resnet0);
	\draw [->] (resnet0) -- (resnet1);
	\draw [->] (resnet1) -- (resnet2);
	\draw [->] (resnet2) -- (resnet3);
	\draw [->] (resnet3) -- node[xshift=0.15cm, yshift=-0.1cm]{\footnotesize $\bm{H}_t$} (6.0, 1.0);		
\end{tikzpicture}
\caption{Memory update based on Residual Units.}
\label{fig:temporalMemory}
\vspace{-0.1cm}
\end{figure}
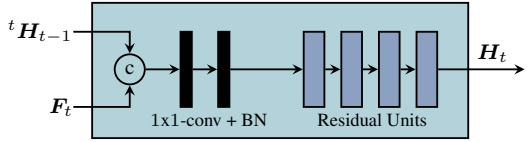

The goal of the proposed architecture is to exploit features and thereby information from previous frames to improve the computed semantic segmentation. Built from three main components, the presented memory alignment, a single frame backbone and a memory module, the recurrent fully convolutional architecture is illustrated in Fig. \ref{fig:fusionArch}. A final 1$\times$1-convolution and softmax layer compute the predictions for every frame.
\vspace{-0.5cm}\paragraph{Single Frame Backbone (SFB)} The point clouds are fed as range images to a backbone network, which computes an intermediate feature map for each individual frame. It is based on deep layer aggregation \cite{Yu2017DeepLA} and closely related to the LaserNet backbone \cite{Meyer2019LaserNetAE}. The only differences to the latter are a reduction of the Residual Units \cite{He2015DeepRL} in the first two feature extractors to 4 and 5 as well as the removal of the downsampling in the first feature extractor.%
\vspace{-0.5cm}\paragraph{Temporal Memory} The key property of our architecture, and RNNs in general, is its recurrent structure, which enables a recursive aggregation of information over time. Not only the current feature map $\bm{F}_t$ is used for the semantic segmentation but also the temporal memory $\bm{H}_{t-1}$ from the previous frames, which contains the recursively aggregated information from the past. We investigate two different memory update strategies, both applied to the temporally aligned memory $\temporalVar{\text{H}}$ and current feature map $\bm{F}_t$ to compute the updated memory $\bm{H}_t$.\\%
\textit{Residual Network}: The first memory update module is based on four Residual Units and illustrated in Fig. \ref{fig:temporalMemory}. It concatenates the temporally aligned memory $\temporalVar{\bm{H}}$ and the intermediate feature map $\bm{F}_t$ from the single frame backbone. Afterwards, an 1$\times$1-convolution layer halves the feature channels followed by a BatchNorm \cite{Ioffe2015BatchNA}. Finally, four Residual Units compute the updated memory $\bm{H}_{t}$. \\% 
Usually, LSTMs \cite{Hochreiter1997} or GRUs \cite{Cho2014} are used to counteract the vanishing or exploding gradient problem in RNNs with their gating mechanisms. The severity of this problem however, depends on the number of time steps the gradient is propagated back, see Fig. \ref{fig:fusionArch} for the gradient flow. For the given segmentation task, we focus on short term dependencies and additionally, memory limitations restrict the number of possible time steps. With a maximum of five, the gradient is backpropagated through $39$ Residual Units at most, which is few more then the ResNet101 architecture, which is easily trainable. \\%
\textit{Gated Recurrent Unit}: The second temporal fusion module, which uses gating mechanisms, is a ConvGRU. GRUs in general require less memory and computational resources than LSTMs while showing similar performance \cite{Chung2014}. A ConvGRU replaces fully connected layers by convolution layers to preserve spatial resolution. With the convolution kernels $\bm{W}_z, \bm{W}_r, \bm{W}$ and $\bm{U}_z, \bm{U}_r, \bm{U}$,  the memory update of GRUs can be described by: 
\begin{gather}
\bm{Z}_t = \sigma \left(\bm{W}_z * \bm{F}_t + \bm{U}_z * \temporalVar{\bm{H}} \right) \\
\bm{R}_t = \sigma \left(\bm{W}_r * \bm{F}_t + \bm{U}_r * \temporalVar{\bm{H}} \right) \\
\bm{H}'_t = \tanh \left(\bm{W} * \bm{F}_t + \bm{U} * \left(\bm{R}_t \circ \temporalVar{\bm{H}} \right) \right) \\
\bm{H}_t = \left(1 - \bm{Z}_t\right) \circ \temporalVar{\bm{H}} + \bm{Z}_t \circ \bm{H}'_t,
\end{gather}%
with convolution operator $*$ and Hadamard product $\circ$.%
%----------------------------------------------------------------------------------------------------------------------------------------------------------------------------------
%----------------------------------------------------------------------------------------------------------------------------------------------------------------------------------
%##########################################################################################################################################
%##########################################################################################################################################
\subsection{Temporal Training}
\begin{figure}[!t]
\centering
\begin{tikzpicture}[line width=0.7pt, >=stealth, auto]
	\definecolor{ncolC}{RGB}{85, 155, 170}
	\definecolor{ncolD}{RGB}{100, 125, 160}  
	\definecolor{ncolE}{RGB}{160, 75, 165}

	% sequence - nodes
	\foreach \x in {6,...,9} 		
		\node[rectangle, font=\footnotesize, fill=ncolC, inner sep=2pt, minimum height=0.5cm, minimum width=0.45cm] at (\x * 0.75, 0.0) {$\x$};
	\node[rectangle, font=\footnotesize, fill=ncolC, draw=ncolE, line width=1.0, inner sep=2pt, minimum height=0.5cm, minimum width=0.45cm] at (7.5, 0.0) {$10$};
	\foreach \x in {11,...,15} 		
		\node[rectangle, font=\footnotesize, fill=ncolD, inner sep=2pt, minimum height=0.5cm, minimum width=0.45cm] at (\x * 0.75, 0.0) {$\x$};
	% sequence - edges
	\draw[->, dotted] (3.8 ,0.0) -> +(0.45,0.0);
	\foreach \x in {6,...,14} 		
		\draw[->] (0.225 + \x * 0.75 ,0.0) -> +(0.3,0.0);
	\draw[->, dotted] (11.5 ,0.0) -> +(0.45,0.0);

	% loss
	\node[rectangle, font=\footnotesize, fill=red, fill opacity=0.4, text opacity=1.0, inner sep=1.25pt, minimum height=0.45cm, minimum width=0.45cm] at (7.5, 0.75) {loss};	
	\draw[->] (7.55 ,0.255) -> +(0.0,0.27);
	\node[rectangle, font=\footnotesize, fill=red, fill opacity=0.4, text opacity=1.0, inner sep=1.25pt, minimum height=0.45cm, minimum width=0.45cm] at (11.25, 0.75) {loss};	
	\draw[->] (11.25 ,0.255) -> +(0.0,0.27);	
	
	% parameter declarations
	\draw[<->, dashed] (7.78 ,0.75) -> +(3.185,0.0) node[font=\footnotesize, align=center, xshift=-1.6cm, yshift=0.15cm] {every $k_1$-th ($=5$) steps};	
	\draw[->, dashed, orange] (7.45 ,0.525) |- +(-3.0,-0.2) node[font=\footnotesize, black, align=center, xshift=1.2cm, yshift=0.35cm] {gradient flow through time \\ for $k_2$ ($=5$) steps};	
	\draw[->, draw=ncolE] (7.5 ,-0.475) -> +(0.0,0.22) node[font=\footnotesize, xshift=0.1cm, yshift=-0.3cm] {first update at $k_3$ ($=10$)};
	
\end{tikzpicture}
\vspace{-0.1cm}
\caption{Parameters $k_1$, $k_2$ and $k_3$, for a sequence of frames numbered by their sequence index, with exemplary values.}
\label{fig:temporalTraining}
\vspace{-0.05cm}
\end{figure}
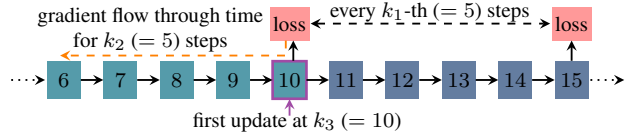

When training recurrent architectures, training and gradient flow differ from training single frame networks. Since the current predictions also depend on previous frames, the error can be backpropagated through time, which is illustrated in Fig. \ref{fig:fusionArch}. Thereby, the single frame backbone learns to compute intermediate feature maps which are not only valuable for the current frame, but also for the next frames. The memory module learns temporal dependencies. \\
We propose a sequence based training for our recurrent architecture, since training with randomly drawn samples from the dataset does not work. Temporal dependencies only exist for a sequence of consecutive frames and therefore, one general prerequisite is, that the underlying dataset provides sequential data. Since short term dependencies are the most valuable, the model is trained with sub-sequences of 25 frames. Thereby, the model is trained with a large variety of sequences in a varying order, more than the dataset itself would provide. The short sub-sequences, which are started with a zero-filled memory, force the network to focus on short term dependencies. To make the backpropagation computational manageable, the backpropagation is truncated  after $k_2$ steps (TBPTT) \cite{Williams1990AnEG} and only performed every $k_1$ steps, see Fig. \ref{fig:temporalTraining}. To allow the memory to aggregate meaningful temporal information before updating, the first update is delayed for at least $k_3$ steps. One general advantage is that not every frame needs to be labeled, a label for every $k_1$-th frame is sufficient.
With given labels for every frame however, the variety of sub-sequences is considerably increased.
%----------------------------------------------------------------------------------------------------------------------------------------------------------------------------------
%----------------------------------------------------------------------------------------------------------------------------------------------------------------------------------
%----------------------------------------------------------------------------------------------------------------------------------------------------------------------------------
%----------------------------------------------------------------------------------------------------------------------------------------------------------------------------------
\section{Experiments}
We evaluated our approach \approach on two challenging datasets. While SemanticKITTI \cite{Behley2019SemanticKITTIAD} provides few but long sequences with a medium number of dynamic objects,
PandaSet provides more but very short sequences with a large number of dynamic objects. More details are described below. For both datasets, the mean Intersection-over-Union (mIoU) is reported as evaluation metric. \\%
\textbf{SemanticKITTI} is a large-scale benchmark with point-wise annotations for $360^{\circ}$ lidar scans of a Velodyne-HDL-64E based on the KITTI Odometry Benchmark \cite{Geiger2012AreWR}. It includes a total of \({22}\) labeled sequences with over \({43,000}\) labeled scans recorded with $10\,\text{Hz}$. The official split allocates sequences \({0}\)--\({10}\) with over ${21,000}$ frames for training and sequences \({11}\)--\({21}\) with no labels published for testing. The single scan benchmark evaluates 19 different classes, the multiple scan benchmark 25. While we follow the official split \cite{milioto2019iros} for the comparison with other state-of-the-art methods, we choose a larger evaluation split for our ablation study. For more significant conclusions, we conduct our ablation study on a validation split consisting of sequences \({02}\), \({06}\) and \({10}\). \\%
\textbf{PandaSet} is a new dataset with point-wise annotations for $360^{\circ}$ lidar scans of a Hesai-Pandar64 and for $60^{\circ}$ front lidar scans of a Hesai-PandarGT. Only the scans of the first sensor are used in the experiments. PandaSet provides $103$ sequences with a length of $80$ frames or $8s$. However, only $76$ sequences have pointwise annotations implying $6080$ labeled frames. No official benchmark or data split has been proposed so far or was established by published work. Hence, we define a train-/val-/test-split with $4320$/$640$/$1120$ frames for our experiments, described in the supplementary material. Trying to match the classes of SemanticKITTI, we group labeled classes together and train on a subset of 14 classes. The grouping or mapping of the labeled classes is described in the supplementary material.%
%----------------------------------------------------------------------------------------------------------------------------------------------------------------------------------
%---------------------------------------------------------------------------------------------------------------------------------------------------------------------------------- 
%##########################################################################################################################################
%##########################################################################################################################################
\subsection{Implementation Details}
All experiments are implemented in PyTorch\footnote{\url{https://pytorch.org/}} and trained in mixed precision mode\footnote{\url{https://github.com/NVIDIA/apex}}. We use distributed data parallel training
on up to 4 Tesla V100 GPUs and synchronize batch norm across GPUs. \\%
Adam optimizer with weight decay of $0.0005$ is used to optimize class-balanced cross entropy loss for $75k$ iterations. The weights are computed by $w_c = \log\left(\nicefrac{n_c}{n}\right)$, where $n_c$ is the number of occurrences of class c and $n$ is the total number of points. To reduce overfitting, we apply random horizontal flip with a probability of 0.5 and use random crops of size $64 \times 1024$ during training. \\%
The \textbf{Single Frame Backbone}, which is also one of the baselines, is trained with different batch sizes. For the pretrained backbone use case, a batch size of $\underline{100}$ is used, but for baseline comparisons, we also investigate the batch size used for \approachq. Thereby, we eliminate any improvements of our approach over the baseline, which are solely induced by a different batch size. Hence, the backbone is additionally trained with a batch size of $20$ for SemanticKITTI as well as $4$ and $\underline{20}$ for PandaSet. The underlined batch sizes achieve the best results and are used as baseline. An initial learning rate of $0.001$ is used, which exponentially decays by $e^{-5\cdot10^{-5}\cdot it}$ with iteration $it$. \\%
Using the pretrained backbone, \textbf{\approachq} with residual memory update is trained with a batch size of $20$ for both datasets and with a batch size of $4$ for the ConvGRU memory update on PandaSet. The approach is trained with an initial learning rate of $0.0001$, which was exponentially decayed by $e^{-2.5\cdot10^{-5}\cdot it}$. It is trained with TBPTT where we choose $k_2 = 5$, limited by memory requirements. By matching $k_1 = k_2$ every frame is considered exactly once for optimization. The first update is performed at frame $k_3 = 10$.%
%----------------------------------------------------------------------------------------------------------------------------------------------------------------------------------
%----------------------------------------------------------------------------------------------------------------------------------------------------------------------------------
\subsection{Recurrent Semantic Segmentation}
Starting with a brief evaluation of our adapted projection strategy, the achieved improvements are shown in Table \ref{tab:ablationProjectionDensitry}. Since the vertical laser distribution of the sensor used in PandaSet is
highly non-uniform, the baseline projection strategy does not work well. With the proposed projection strategy however, about $90\%$ of the points are projected collision-free into the range image, in contrast to less than half. For SemanticKITTI and its sensor, the improvements are much smaller, because of its nearly uniform laser distribution. In this case, we prefer the simple strategy because it is faster.  \\%
The main goal of the presented approach is the exploitation of temporal dependencies to improve segmentation results. The influence of our individual components, and therefore the results of an increasing exploitation of temporal information, is illustrated in Table \ref{tab:ablationComponents}. An initial improvement is achieved when adding a temporal memory, which provides temporal information from the past. The memory however is unaligned and because of the missing TBPTT, has considerable limited capabilities of learning temporal dependencies. Adding TMA further improves the segmentation while TBPTT only improves the results if combined with TMA. This underlines the importance of the proposed alignment strategy. The best results are therefore achieved with all components and when relying on as much temporal information as possible. Considering computational complexity, only the addition of the temporal memory increases runtime, TMA can be computed in parallel to the SFB and TBPTT only affects training. Reported runtimes are mean values over sequence 08 of SemanticKITTI. \\%
We investigate the presented improvements in more depth by conducting three additional experiments. First, we train our single frame backbone with additional four Residual Units. The results are shown in Table \ref{tab:ablationTemporalInfluence} and verify that not the increased model capacity but the learned temporal dependencies improve the results. Second, we evaluate \approach with the temporal memory set to zero. The considerably worse results, shown in Table \ref{tab:ablationTemporalInfluence}, emphasize that the presented approach learns and strongly relies on temporal dependencies. Finally, we consider a simple aggregation strategy,
aggregating the predictions of the last five frames by majority voting. The results in Table \ref{tab:ablationTemporalInfluence} underline that a more complex approach is required to achieve the presented improvements. \\%
%
%----------------------------------------------------------------------------------------------------------------------------------------------------------------------------------
%----------------------------------------------------------------------------------------------------------------------------------------------------------------------------------
%##########################################################################################################################################
\begin{table}
\centering
\renewcommand{\arraystretch}{1.2}
\begin{tabular}{ l | c | c}
\hline
& baseline & \textit{Ours}\\
\hline \hline
PandaSet & $43.3$ & $\bm{89.3}$ \\
SemanticKITTI & $77.2$ & $\bm{81.5}$ \\
\hline
\end{tabular}%
\caption{Mean fraction of points, which are projected collision-free into the range image. Values are in $\%$.}%
\label{tab:ablationProjectionDensitry}%
\end{table}%
%##########################################################################################################################################
%##########################################################################################################################################
\begin{table}
\centering
\renewcommand{\arraystretch}{1.2}
\resizebox{.47\textwidth}{!}{%
\begin{tabular}{ c  c  c  c | c | c | c}
\hline
SFB & TM & TBPTT & TMA & mIoU ($\%$) & runtime & params\\
\hline \hline
\checkmark & & & & $53.5$ & $25.3$ms & $4.25\,$M \\
\hline
\checkmark & \checkmark & & & $55.0$ & \multirow{ 3}{*}[-0.6em]{$33.2$ms} & \multirow{ 3}{*}[-0.6em]{$5.61\,$M} \\
\checkmark & \checkmark & \checkmark & & $55.1$ & &\\
\checkmark & \checkmark & & \checkmark & $56.3$ & & \\
\checkmark & \checkmark & \checkmark & \checkmark & $\bm{57.1}$ & &\\
\hline
\end{tabular}}%
\caption{Improvements, when using a temporal memory or recurrent architecture (TM), training with TBPTT and temporally aligning the memory (TMA).}%
\label{tab:ablationComponents}%
\end{table}%
%##########################################################################################################################################
%##########################################################################################################################################
\begin{table}
\centering
\renewcommand{\arraystretch}{1.2}
\begin{tabular}{ l | c }
\hline
Approach & mIoU ($\%$)\\
\hline \hline
SFB + 4 Residual Units & $53.9$\\
\approachq, empty memory  & $52.4$\\
Aggregation by majority voting & $54.4$\\
\hline
\approach & $\bm{57.1}$\\
\hline
\end{tabular}%
\caption{Importance of temporal information.}%
\label{tab:ablationTemporalInfluence}%
\end{table}%
%##########################################################################################################################################
%##########################################################################################################################################
\begin{table*}%
\centering%
\addtolength{\tabcolsep}{-2.5pt}%
\def\arraystretch{1.25}%   
\resizebox{.8\textwidth}{!}{%
\begin{tabular}{ l | c | c c c c c c c c c c c c c c }
  \hline 
Approach & mIoU & \rot90{car } & \rot90{bicycle } & \rot90{motorcycle } & \rot90{truck } & \rot90{other-vehicle } & \rot90{person } & \rot90{road } & \rot90{road barriers } & \rot90{sidewalk } & \rot90{building } & \rot90{vegetation } & \rot90{terrain } & \rot90{background } & \rot90{traffic sign }  \\
  \hline \hline
  SqueezeSegV3 \cite{Xu2020} & $55.7$ & $92.8$ & $24.1$ & $18.0$ & $36.5$ & $54.3$ & $63.0$ & $\bm{91.1}$ & $11.9$ & $\bm{71.3}$ & $86.2$ & $85.0$ & $61.3$ & $63.2$ & $20.6$ \\
  SalsaNext \cite{Cortinhal2020} & $57.8$ & $92.1$ & $\bm{40.7}$ & $31.7$ & $28.7$ & $56.2$ & $69.0$ & $90.0$ & $\bm{22.6}$ & $67.1$ & $85.6$ & $83.4$ & $58.5$ & $63.3$ & $20.6$ \\
  \hline \hline 
  SFB (\textit{Ours}) & $58.3$ & $93.0$ & $32.8$ & $30.2$ & $34.6$ & $55.9$ & $69.4$ & $90.9$ & $14.0$ & $68.7$ & $87.1$ & $87.0$ & $60.5$ & $66.3$ & $\bm{26.0}$ \\
  \approach (\textit{Ours}) & $\bm{60.0}$ & $\bm{93.7}$ & $33.6$ & $38.0$ & $37.1$ & $\bm{59.9}$ & $\bm{72.0}$ & $\bm{91.1}$ & $14.6$ & $70.6$ & $\bm{88.2}$ & $\bm{88.4}$ & $\bm{63.8}$ & $\bm{68.4}$ & $20.7$ \\
  \hline
  \approachq-GRU (\textit{Ours}) & $59.7$ & $93.5$ & $35.5$ & $\bm{43.8}$ & $\bm{39.3}$ & $54.8$ & $70.8$ & $90.6$ & $12.2$ & $69.1$ & $87.4$ & $87.2$ & $63.5$ & $66.4$ & $22.2$ \\
  \hline  
\end{tabular}}%
\addtolength{\tabcolsep}{1pt}%
\caption{Results on the test split of PandaSet. Values are given as IoU ($\%$).}%
\label{tab:metricsPanda}%
\end{table*}%
%##########################################################################################################################################
%##########################################################################################################################################
\begin{table*}%
\centering%
\addtolength{\tabcolsep}{-2.5pt}%
\def\arraystretch{1.25}%   
\resizebox{\textwidth}{!}{%
\begin{tabular}{ l | c | c c c c c c c c c c c c c c c c c c c c c c c c c}
  \hline
Approach & mIoU & \rot90{car } & \rot90{bicycle } & \rot90{motorcycle } & \rot90{truck } & \rot90{other-vehicle } & \rot90{person } & \rot90{bicyclist } & \rot90{motorcyclist } & \rot90{road } & \rot90{parking } & \rot90{sidewalk } & \rot90{other-ground } & \rot90{building } & \rot90{fence } & \rot90{vegetation } & \rot90{trunk } & \rot90{terrain } & \rot90{pole } & \rot90{traffic sign } & \rot90{$\underrightarrow{\text{car}}$ } & \rot90{$\underrightarrow{\text{bicyclist}}$ } & \rot90{$\underrightarrow{\text{person}}$ } & \rot90{$\underrightarrow{\text{motorcyclist}}$t } & \rot90{$\underrightarrow{\text{other-vehicle}}$ } & \rot90{$\underrightarrow{\text{truck}}$ } \\
  \hline \hline
  TangentConv \cite{Tatarchenko2018} & $34.1$ & $84.9$ & $2.0$ & $18.2$ & $21.1$ & $18.5$ & $1.6$ & $0.0$ & $0.0$ & $83.9$ & $38.3$ & $64.0$ & $15.3$ & $85.8$ & $49.1$ & $79.5$ & $43.2$ & $56.7$ & $36.4$ & $31.2$ & $40.3$ & $1.1$ & $6.4$ & $1.9$ & $\bm{30.1}$ & $\bm{42.2}$ \\
  DarkNet53Seg \cite{Behley2019SemanticKITTIAD} & $41.6$ & $84.1$ & $30.4$ & $32.9$ & $20.2$ & $20.7$ & $7.5$ & $0.0$ & $0.0$ & $91.6$ & $\bm{64.9}$ & $75.3$ & $\bm{27.5}$ & $85.2$ & $56.5$ & $78.4$ & $50.7$ & $64.8$ & $38.$1 & $53.3$ & $61.5$ & $14.1$ & $15.2$ & $0.2$ & $28.9$ & $37.8$ \\
  SpSequenceNet \cite{ShiSpSequenceNetSS} & $43.1$ & $88.5$ & $24.0$ & $26.2$ & $29.2$ & $22.7$ & $6.3$ & $0.0$ & $0.0$ & $90.1$ & $57.6$ & $73.9$ & $27.1$ & $\bm{91.2}$ & $\bm{66.8}$ & $\bm{84.0}$ & $\bm{66.0}$ & $\bm{65.7}$ & $50.8$ & $48.7$ & $53.2$ & $41.2$ & $26.2$ & $\bm{36.2}$ & $2.3$ & $0.1$ \\
  \hline 
  \approach (\textit{Ours}) & $\bm{47.0}$ & $\bm{92.1}$ & $\bm{47.7}$ & $\bm{40.9}$ & $\bm{39.2}$ & $\bm{35.0}$ & $\bm{14.4}$ & $0.0$ & $0.0$ & $\bm{91.8}$ & $59.6$ & $\bm{75.8}$ & $23.2$ & $89.8$ & $63.8$ & $82.3$ & $62.5$ & $64.7$ & $\bm{52.6}$ & $\bm{60.4}$ & $\bm{68.2}$ & $\bm{42.8}$ & $\bm{40.4}$ & $12.9$ & $12.4$ & $2.1$ \\
  \hline  
\end{tabular}}%
\addtolength{\tabcolsep}{1pt}  %
\caption{Comparison to state-of-the-art methods on the official SemanticKITTI multiple scans benchmark. The arrow below classes indicate moving classes. Values are given as IoU ($\%$).}%
\label{tab:metricsKittiDyn}%
\end{table*}%
%##########################################################################################################################################
%
While mainly building on the residual memory update, we also investigate a ConvGRU for the memory update. Therefore, we train our approach with a ConvGRU on PandaSet, with the results illustrated in Table \ref{tab:metricsPanda}. Independently from the chosen memory update, \approach improves the results compared to SFB, the residual memory update however, performs better.
\subsection{Quantitative Benchmark Results}
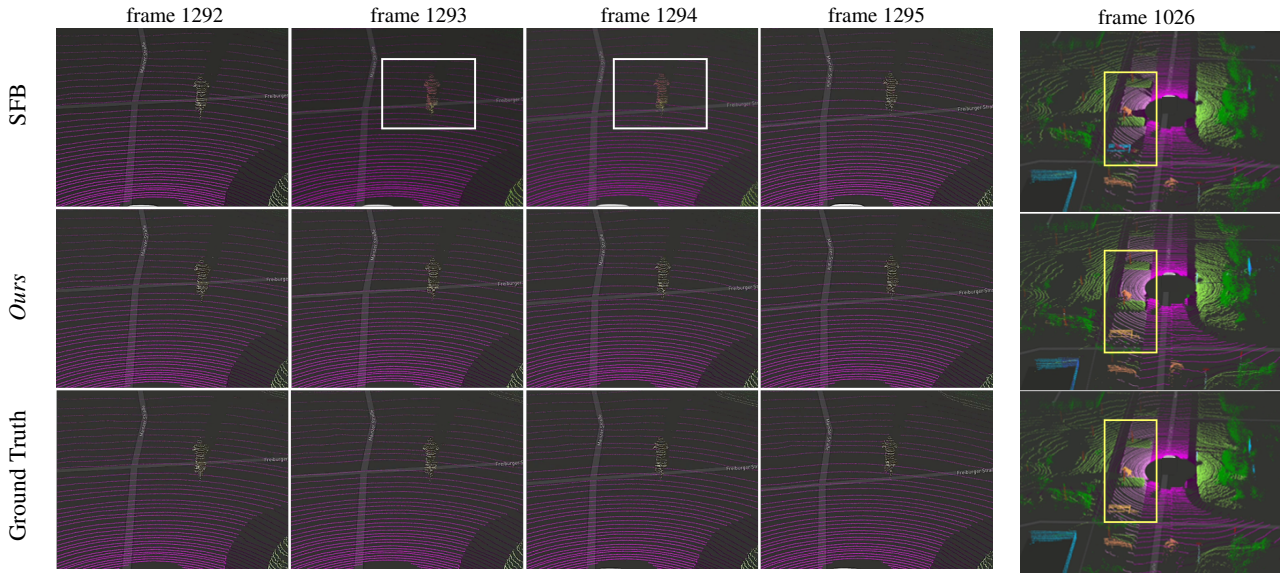
\begin{figure*}[t]
\centering
\begin{subfigure}{.75\textwidth}
  \centering
  \begin{overpic}[width=0.95\linewidth]{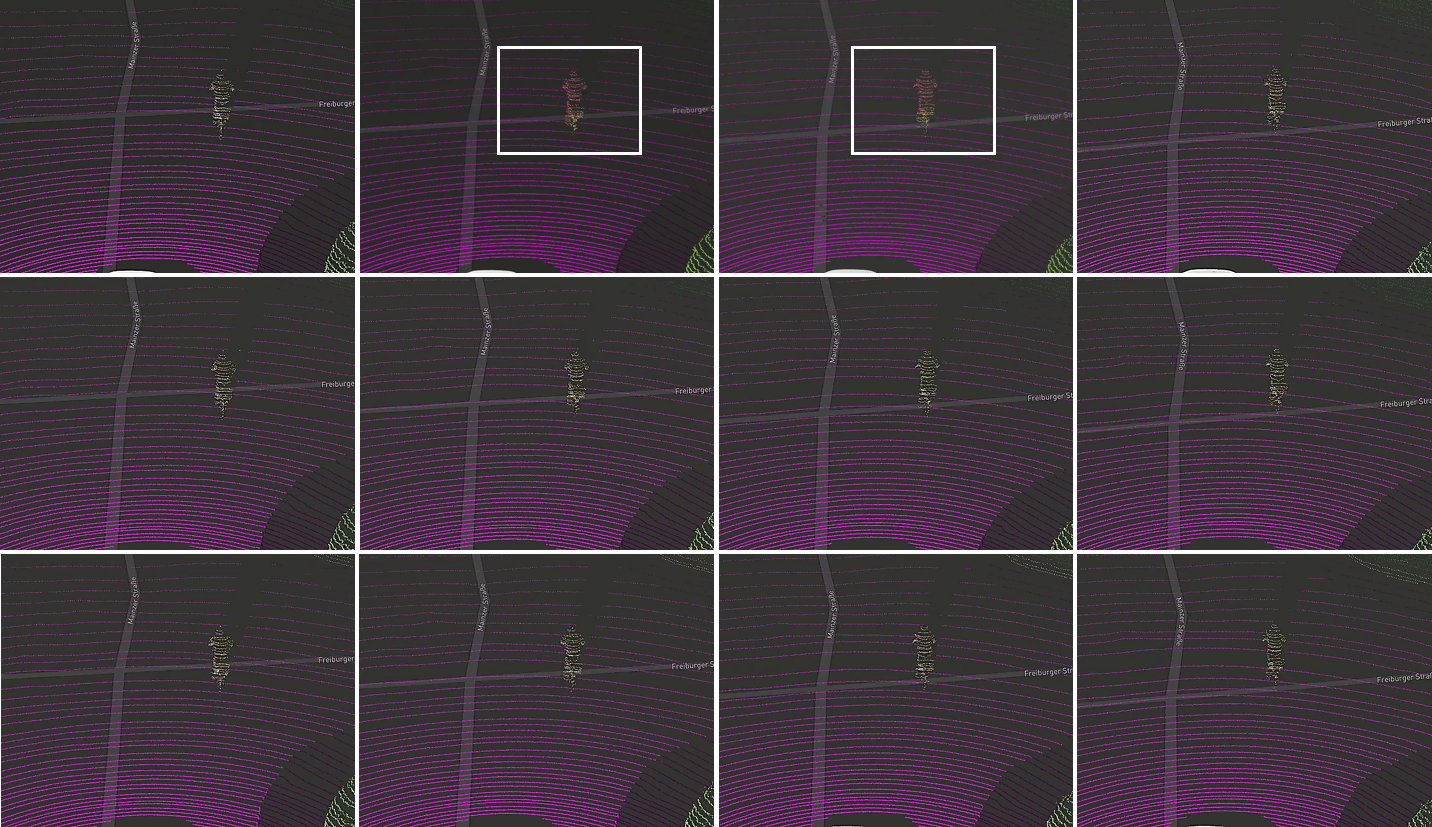}
  \put(-5, 47.4){\small \rotatebox{90}{SFB}}
  \put(-5, 26.5){\small \textit{\rotatebox{90}{Ours}}}
  \put(-5, 3){\small \rotatebox{90}{Ground Truth}}
  \put(7.5, 58.5){\footnotesize frame 1292}
  \put(33.6, 58.5){\footnotesize frame 1293}
  \put(58.2, 58.5){\footnotesize frame 1294}
  \put(82.5, 58.5){\footnotesize frame 1295}
  \end{overpic}  
  \label{fig:sub1}
\end{subfigure}%
\begin{subfigure}{.2\textwidth}
  \centering
  \begin{overpic}[width=0.985\linewidth]{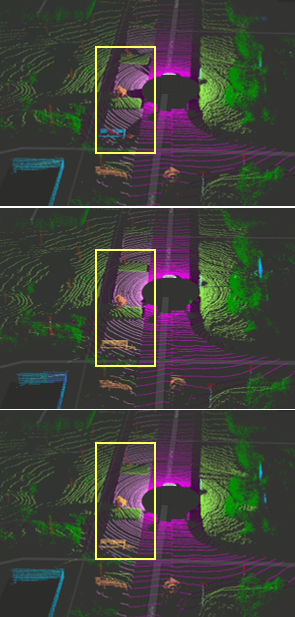}
    \put(14.3, 101.5){\footnotesize frame 1026}
   \end{overpic}
  \label{fig:sub2}
\end{subfigure}
\caption{Qualitative experiments on validation sequence 08. While SFB switches between bicycle \protect\colBox{col_bicycle} and person \protect\colBox{col_person} over the frames, \protect\approach is more stable and correctly classifies the bicycle in all frames. On the right, in a different scene, \approach accurately segments the whole parking area \protect\colBox{col_parking}, SFB however confuses some of the area with sidewalk \protect\colBox{col_sidewalk}. Our approach is also able to segment the trailer \protect\colBox{col_other_veh} while SFB classifies the points as building \protect\colBox{col_building}.}
\label{fig:qualitativeResults}
\end{figure*}

We evaluate the proposed recurrent segmentation architecture on the multiple scans benchmark of SemanticKITTI, which targets the exploitation of past scans to improve segmentation results. One important additional property is the requirement to distinguish moving from non moving for dynamic classes. This requires a model to consider temporal dependencies, to understand the movement of objects. The results for this challenging task are illustrated in Table \ref{tab:metricsKittiDyn}. \approachq\footnote{name of our submissions: sem\_seg\_3d} outperforms all existing approaches by a considerable margin. Especially for most of the dynamic
classes our approach performs significantly better. This underlines the capabilities of our model to learn and exploit temporal dependencies. The improvements for moving dynamic classes require a deeper investigation. In general, a systematic error is introduced by TMA for moving object, because their motion is not considered. Objects moving fast and tangential to a circle around and near the ego vehicle cause the largest errors. Because the error diminishes with distance, non-tangential movement direction and slower speed, this error is rather small for most objects. On average, the value of the temporal dependencies therefore outweigh the spatial error.  Looking at the static classes \approach performs slightly better for some classes,  while \cite{ShiSpSequenceNetSS} performs slightly better for others. For traffic signs however, our approach considerably outperforms all others again.\\%
In the next step, we investigate the improvements compared to approaches, which do not use temporal information. Therefore, \approach and SFB are evaluated on the SemanticKITTI single scan benchmark. Their comparison, shown in Table \ref{tab:metricsKitti}, reveals the benefits of our proposed approach. Overall, our approach performs significantly better, outperforming SFB on every class. \approach shows that dynamic as well as static classes benefit from the exploitation of temporal dependencies. Qualitative results are illustrated in Fig. \ref{fig:qualitativeResults}. Compared to other state-of-the-art approaches, our approach outperforms every projection based method by a considerable margin. One exception is SalsaNext trained with Lovász-Softmax loss \cite{Berman2017}. While we also outperform  SalsaNext when trained with uncertainty, Lovász-Softmax loss greatly increased their performance (mIoU +6.5). Considering also point based methods, only KPConv is slightly better than ours, but also much slower.  \\%
%##########################################################################################################################################
\begin{table*}%
\centering%
\addtolength{\tabcolsep}{-2.5pt}%
\def\arraystretch{1.25}%   
\resizebox{\textwidth}{!}{%
\begin{tabular}{ l | c | c c c c c c c c c c c c c c c c c c c }
  \hline
Approach & mIoU & \rot90{car } & \rot90{bicycle } & \rot90{motorcycle } & \rot90{truck } & \rot90{other-vehicle } & \rot90{person } & \rot90{bicyclist } & \rot90{motorcyclist } & \rot90{road } & \rot90{parking } & \rot90{sidewalk } & \rot90{other-ground } & \rot90{building } & \rot90{fence } & \rot90{vegetation } & \rot90{trunk } & \rot90{terrain } & \rot90{pole } & \rot90{traffic sign }  \\
  \hline \hline
  LatticeNet \cite{Rosu2019} & $52.2$ & $88.6$ & $12.0$ & $20.8$ & $\bm{43.3}$ & $24.8$ & $34.2$ & $39.9$ & $\bm{60.9}$ & $88.8$ & $64.6$ & $73.8$ & $25.5$ & $86.9$ & $55.2$ & $76.4$ & $67.9$ & $54.7$ & $41.5$ & $42.7$ \\
  RandLA-Net \cite{Hu2019} & $53.9$ & $94.2$ & $26.0$ & $25.8$ & $40.1$ & $38.9$ & $49.2$ & $48.2$ & $7.2$ & $90.7$ & $60.3$ & $73.7$ & $20.4$ & $86.9$ & $56.3$ & $81.4$ & $61.3$ & $66.8$ & $49.2$ & $47.7$ \\
  KPConv \cite{Thomas2019} & $58.8$ & $\bm{96.0}$ & $30.2$ & $42.5$ & $33.4$ & $\bm{44.3}$ & $\bm{61.5}$ & $\bm{61.6}$ & $11.8$ & $88.8$ & $61.3$ & $72.7$ & $\bm{31.6}$ & $\bm{90.5}$ & $\bm{64.2}$ & $\bm{84.8}$ & $\bm{69.2}$ & $\bm{69.1}$ & $\bm{56.4}$ & $47.4$ \\
  \hline 
  RangeNet53++ \cite{milioto2019iros} & $52.5$ & $91.4$ & $25.7$ & $34.4$ & $25.7$ & $23.0$ & $38.3$ & $38.8$ & $4.8$ & $91.8$ & $\bm{65.0}$ & $75.2$ & $27.8$ & $87.4$ & $58.6$ & $80.5$ & $55.1$ & $64.6$ & $47.9$ & $55.9$ \\
  PolarNet \cite{Zhang2020} & $54.3$ & $83.8$ & $40.3$ & $30.1$ & $22.9$ & $28.5$ & $43.2$ & $40.2$ & $5.6$ & $90.8$ & $61.7$ & $74.4$ & $21.7$ & $90.0$ & $61.3$ & $84.0$ & $65.5$ & $67.8$ & $51.8$ & $57.5$ \\
  3DMiniNet++ \cite{Alonso2020} & $55.8$ & $90.5$ & $42.3$ & $42.1$ & $28.5$ & $29.4$ & $47.8$ & $44.1$ & $14.5$ & $\bm{91.9}$ & $64.2$ & $74.5$ & $25.4$ & $89.4$ & $60.8$ & $82.8$ & $60.8$ & $66.7$ & $48.0$ & $56.6$ \\
  SqueezeSegV3 \cite{Xu2020} & $55.9$ & $92.5$ & $38.7$ & $36.5$ & $29.6$ & $33.0$ & $45.6$ & $46.2$ & $20.1$ & $91.7$ & $63.4$ & $74.8$ & $26.4$ & $89.0$ & $59.4$ & $82.0$ & $58.7$ & $65.4$ & $49.6$ & $58.9$ \\
  SalsaNext+Uncert \cite{Cortinhal2020} & $55.8$ & $91.6$ & $40.7$ & $26.0$ & $28.2$ & $24.4$ & $53.7$ & $54.1$ & $12.1$ & $91.7$ & $63.1$ & $74.9$ & $25.1$ & $90.4$ & $62.5$ & $82.3$ & $64.0$ & $66.5$ & $53.5$ & $56.1$ \\
  SalsaNext \cite{Cortinhal2020} & $\bm{59.5}$ & $91.9$ & $48.3$ & $38.6$ & $38.9$ & $31.9$ & $60.2$ & $59.0$ & $19.4$ & $91.7$ & $63.7$ & $75.8$ & $29.1$ & $90.2$ & $\bm{64.2}$ & $81.8$ & $63.6$ & $66.5$ & $54.3$ & $\bm{62.1}$ \\
  \hline 
  SFB (\textit{Ours}) & $54.1$ & $91.7$ & $47.9$ & $40.0$ & $19.5$ & $24.7$ & $50.0$ & $42.1$ & $7.3$ & $90.3$ & $53.7$ & $74.3$ & $26.3$ & $86.0$ & $58.4$ & $82.9$ & $61.6$ & $65.7$ & $50.0$ & $55.7$ \\
  \approach (\textit{Ours}) & $58.2$ & $94.1$ & $\bm{50.0}$ & $\bm{45.7}$ & $28.1$ & $37.1$ & $56.8$ & $47.3$ & $9.2$ & $91.7$ & $60.1$ & $\bm{75.9}$ & $27.0$ & $89.4$ & $63.3$ & $83.9$ & $64.6$ & $66.8$ & $53.6$ & $60.5$ \\
  \hline  
\end{tabular}}%
\addtolength{\tabcolsep}{1pt}  %
\caption{Comparison to state-of-the-art methods and to SFB on the official SemanticKITTI single scan benchmark. Values are given as IoU ($\%$).}%
\label{tab:metricsKitti}%
\vspace{0.3cm}
\end{table*}%
%########################################################################################################################################## 
%##########################################################################################################################################
%##########################################################################################################################################
%
Finally, we consider with PandaSet a second data set using also a different sensor. Table \ref{tab:metricsPanda} shows the considerable improvements achieved by \approach compared to SFB. The latter is outperformed for every single class, except traffic sign. As a result, similar to SemanticKITTI, static and dynamic classes benefit from temporal information. In general, the improvements are smaller than for SemanticKitti. One reason are the very short sequences. Therefore, a considerable amount of frames are processed with only few aggregated temporal information in the memory. For further comparison, we also trained two state-of-the-art methods on PandaSet by using their published code and hyper parameters, with the result also shown in Table \ref{tab:metricsPanda}.
%----------------------------------------------------------------------------------------------------------------------------------------------------------------------------------
%----------------------------------------------------------------------------------------------------------------------------------------------------------------------------------
%
\section{Conclusion}
In this work, we proposed a recurrent segmentation architecture using range images as input to improve semantic segmentation of 3d point clouds. The presented approach reuses recursively aggregate and temporally aligned features of past frames to exploit short term temporal dependencies. We demonstrated the benefits of our approach and the value of temporal information for semantic segmentation on three benchmarks of two datasets.
While our approach ranks first on the SemanticKITTI multiple scans benchmark, it also outperforms most of the state-of-the-art methods on the single scan benchmark. Experiments on a second dataset, PandaSet, underlines the benefits. Overall, temporal information provide a huge potential to improve 3d semantic segmentation and thereby enhance the 3d scene understanding of autonomous vehicles and robots.

\FloatBarrier

\pagebreak
{\small
\bibliographystyle{ieee}
\bibliography{egbib}
}
 
\clearpage

\renewcommand*\appendixpagename{Appendix}
\renewcommand*\appendixtocname{Appendix}

\appendix
\appendixpage
\addappheadtotoc
\vspace{0.2cm}
In this section, additional information about the data split and labels used for the evaluation on \textbf{PandaSet} are provided.
We use the following data split:
\begin{itemize}
\item \textbf{Test sequences} \\ 003, 019, 023, 039, 044, 054, 069, 073, 088, 097, 102, 112, 122, 158 
\item \textbf{Validation sequences} \\ 005, 021, 040, 067, 078, 110, 124
\item \textbf{Training sequences} \\ Remaining sequences with point-wise labels provided.
\end{itemize}
The classes used for training are shown in Table \ref{tab:pandaClasses}. They are motivated by the classes of SemanticKITTI with some classes 
missing because PandaSet doesn't provide labels for e.g. parking, fence, trunk or pole. The mapping from the original classes of PandaSet to
the training classes can be found in Table \ref{tab:pandaMapping}. Some of the classes are also ignored during training.
\vspace{1cm}
\begin{table}[h!]
\centering
\renewcommand{\arraystretch}{1.25}
\begin{tabular}{ l | c }
Class & ID \\
\hline
\hline
		car & 0 \\
		bicycle & 1 \\
		motorcycle & 2 \\
		truck & 3 \\
		other-vehicle &	4 \\
		person & 5 \\
		road & 6 \\
		road barriers & 7 \\
		sidewalk & 8 \\
		building & 9 \\
		vegetation & 10 \\
		terrain & 11 \\
		background & 12 \\
		traffic-sign & 13 \\
		ignore & 14 \\
\hline
\end{tabular}%
\caption{Custom classes of PandaSet used for training.}%
\label{tab:pandaClasses}%
\end{table}%
\begin{table}[h!]
\centering
\renewcommand{\arraystretch}{1.2}
\begin{tabular}{l | c | c}
Class & Orig. ID & New ID\\
\hline
\hline
	Smoke & 1 & 14 \\
	Exhaust & 2 & 14 \\
	Spray or Rain & 3 & 14 \\
	Reflection & 4 & 14 \\
	Vegetation & 5 & 10 \\
	Ground & 6 & 11 \\
	Road & 7 & 6 \\
	Lane Line Marking &	8 & 6 \\
	Stop Line Marking &	9 & 6 \\
	Other Road Marking & 10 & 6 \\
	Sidewalk & 11 & 8 \\
	Driveway & 12 & 6 \\
	Car & 13 & 0 \\
	Pickup Truck & 14 & 0 \\
	Medium-sized Truck & 15 & 3 \\
	Semi-truck & 16 & 3 \\
	Towed Object & 17 & 4 \\
	Motorcycle & 18 & 2 \\
	Other Vehicle - Constr. Vehicle & 19 & 4 \\
	Other Vehicle - Uncommon & 20 & 4 \\
	Other Vehicle - Pedicab & 21 & 4 \\
	Emergency Vehicle &	22 & 4 \\
	Bus & 23 & 4 \\
	Personal Mobility Device & 24 & 14 \\
	Motorized Scooter & 25 & 1 \\
	Bicycle & 26 & 1 \\
	Train & 27 & 4 \\
	Trolley & 28 & 4 \\
	Tram / Subway & 29 & 4 \\
	Pedestrian & 30 & 5 \\
	Pedestrian with Object & 31 & 5 \\
	Animals - Bird & 32 & 14 \\
	Animals - Other & 33 & 14 \\
	Pylons & 34 & 7 \\
	Road Barriers & 35 & 7 \\
	Signs & 36 & 13 \\
	Cones &	37 & 7 \\
	Construction Signs & 38 & 13 \\
	Temporary Construction Barriers & 39 & 7 \\
	Rolling Containers & 40 & 12 \\
	Building & 41 & 9 \\
	Other Static Object & 42 & 12 \\
\hline
\end{tabular}%
\caption{Mapping to and grouping in custom classes.}%
\label{tab:pandaMapping}%
\end{table}%

\end{document}